\documentclass[letterpaper, 10 pt, conference]{ieeeconf}  

\IEEEoverridecommandlockouts                              
\usepackage{cite}
\usepackage{amsmath,amssymb,amsfonts}
\usepackage{algorithmic}
\usepackage{graphicx}
\usepackage{textcomp}
\usepackage{xcolor}
\usepackage[lined,boxed,ruled]{algorithm2e}
\usepackage{xcolor, soul}
\sethlcolor{yellow}
\usepackage{booktabs}
\usepackage{adjustbox}
\usepackage{caption}
\usepackage{subcaption}
\usepackage{lipsum}
\usepackage{bm}

\newcommand{\tm}[1]{\text{#1}}
\newcommand\bs[1]{\boldsymbol{#1}}
\DeclareMathOperator*{\argmax}{arg\,max}
\DeclareMathOperator*{\argmin}{arg\,min}

\usepackage{pgfplots}
\usepackage{tikz}
\usetikzlibrary{decorations.pathreplacing,
	backgrounds,
	fit,
	shapes.geometric,
	arrows.meta,
	petri,
	positioning}
\pgfplotsset{compat=1.16}

\makeatletter
\let\NAT@parse\undefined
\makeatother

\usepackage[bookmarks=true]{hyperref}
\definecolor{citationBlue}{RGB}{0,102,204}
\definecolor{linkRed}{RGB}{204,51,0}

\title{\LARGE \bf
	Improving behavior profile discovery for vehicles
}

\author{Nelson de Moura$^{1}$, Fernando Garrido$^{2}$ and Fawzi Nashashibi$^{1}$
	\thanks{$^{1}$Nelson de Moura and Fawzi Nashashibi are with INRIA, 75012 Paris, France,
		{\tt\small \{nelson.demoura;~fawzi.nashashibi\}@inria.fr}}%
	\thanks{$^{2}$Fernando Garrido is with Valeo DSW team, 94000 Créteil, France,
		{\tt\small fernando.garrido@valeo.com}}%
	\thanks{This research has been funded by the plan "France Relance", grant agreement number ANR-21-PRRD-0005-01}		
}

\begin{document}
	
	\maketitle
	\thispagestyle{empty}
	\pagestyle{empty}

	\begin{abstract}
		
		Multiple approaches have already been proposed to mimic real driver behaviors in simulation. This article proposes a new one, based solely on the exploration of undisturbed observation of intersections. From them, the behavior profiles for each macro-maneuver will be discovered. Using the macro-maneuvers already identified in previous works, a comparison method between trajectories with different lengths using an Extended Kalman Filter (EKF) is proposed, which combined with an Expectation-Maximization (EM) inspired method, defines the different clusters that represent the behaviors observed. This is also paired with a Kullback-Liebler divergent (KL) criteria to define when the clusters need to be split or merged. Finally, the behaviors for each macro-maneuver are determined by each cluster discovered, without using any map information about the environment and being dynamically consistent with vehicle motion. By observation it becomes clear that the two main factors for driver's behavior are their assertiveness and interaction with other road users. Paper presented at IROS2024; please use the IEEE citation!
		
	\end{abstract}

	\section{INTRODUCTION}
	
	To predict and model interactions of road users during the driving task is still a challenging necessity for automated vehicles (AVs). Different reasons can influence road users' behavior, producing trajectories that are ultimately distinct from each other. Since the DARPA challenges and the resurgence of AVs as a real technology, multiple methods have been proposed to model road user behaviors and thus make the AV more aware and responsive during driving. Starting from the raw data obtained by the observation of undisturbed intersections, an approach to discover vehicle behaviors in urban conditions is proposed in this work: to extract all differentiable behavior from real data using the longitudinal dynamic information (velocity and acceleration) as the main discriminator.
	
	Given that a sufficient amount of trajectory samples have been observed, the behaviors detected from clustering the set of samples from a single maneuver should represent the totality of the probable behaviors in a specific scenario. They can be used to increase the accuracy and reduce complexity during the prediction of vehicle behavior \cite{bahari2021injecting}, to compare if an AV generated trajectory is similar to a real one (to avoid trajectories that might startle other drivers) or to power the simulation of agents that behave and interact in a realistic manner during simulation \cite{hu2022review}. This last one is the main motivation for the proposed work. For other fields, the general idea of the proposed method can also be useful for the dynamic analysis of human movement during a specific task, or even how different the movements of tools used/hold by surgeons are. 
	
	A particular advantage of the presented method is the discovery of every interaction mode for a maneuver, i.e., every behavior produced by a driver due to a response towards other road users (or the absence of interaction as well) is represented by one of the clusters identified. This is an extremely valuable information, obtained without any kind of hypothesis about how or why the agents interacted with others.
	
	\begin{figure}[!t]
		\centering
		
		\begin{subfigure}[b]{0.39\columnwidth}
			\centering
			\adjustbox{scale=1.5, trim=1mm 0mm 18mm 1mm, clip}{
				\includegraphics[width=1\textwidth]{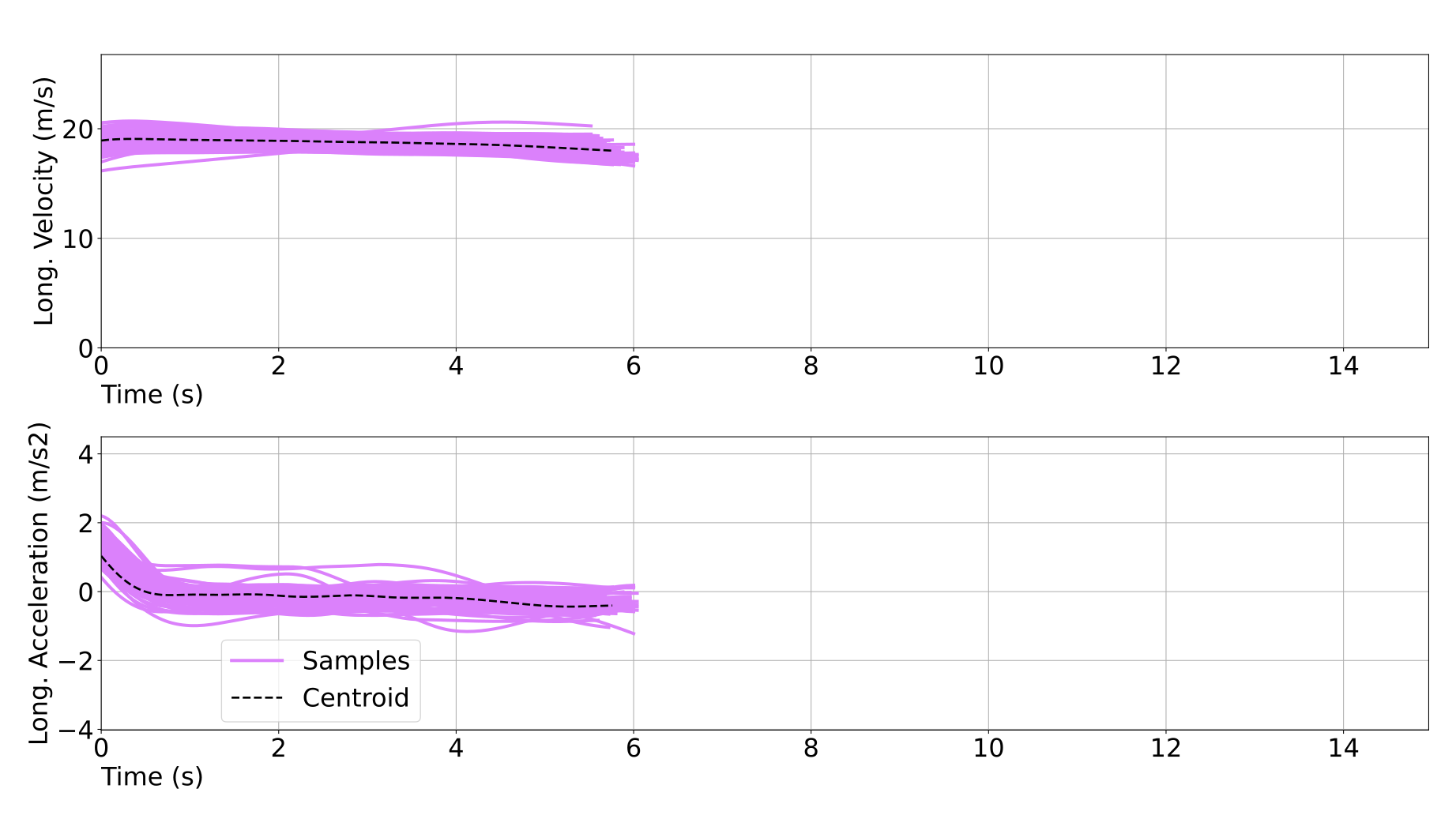} 
			}
		\end{subfigure}
		\begin{subfigure}[b]{0.59\columnwidth}
			\centering
			\adjustbox{scale=1, trim=4.5mm 0mm 0mm 1mm, clip}{
				\includegraphics[width=1\textwidth]{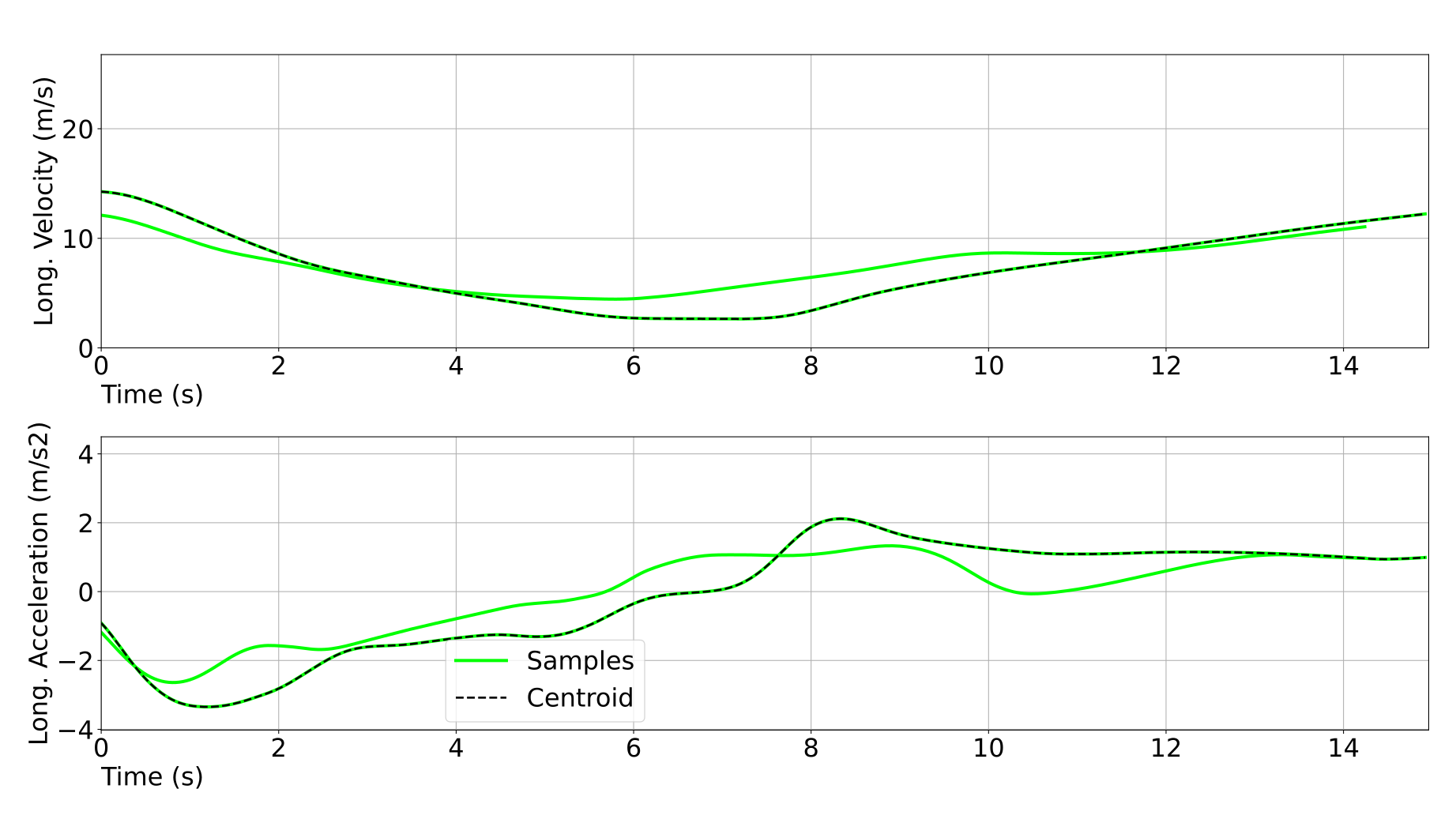} 
			}
		\end{subfigure}

		\begin{subfigure}[b]{0.39\columnwidth}
			\centering
			\adjustbox{scale=1.1, trim=1mm 0mm 7mm 1.25mm, clip}{
				\includegraphics[width=1\textwidth]{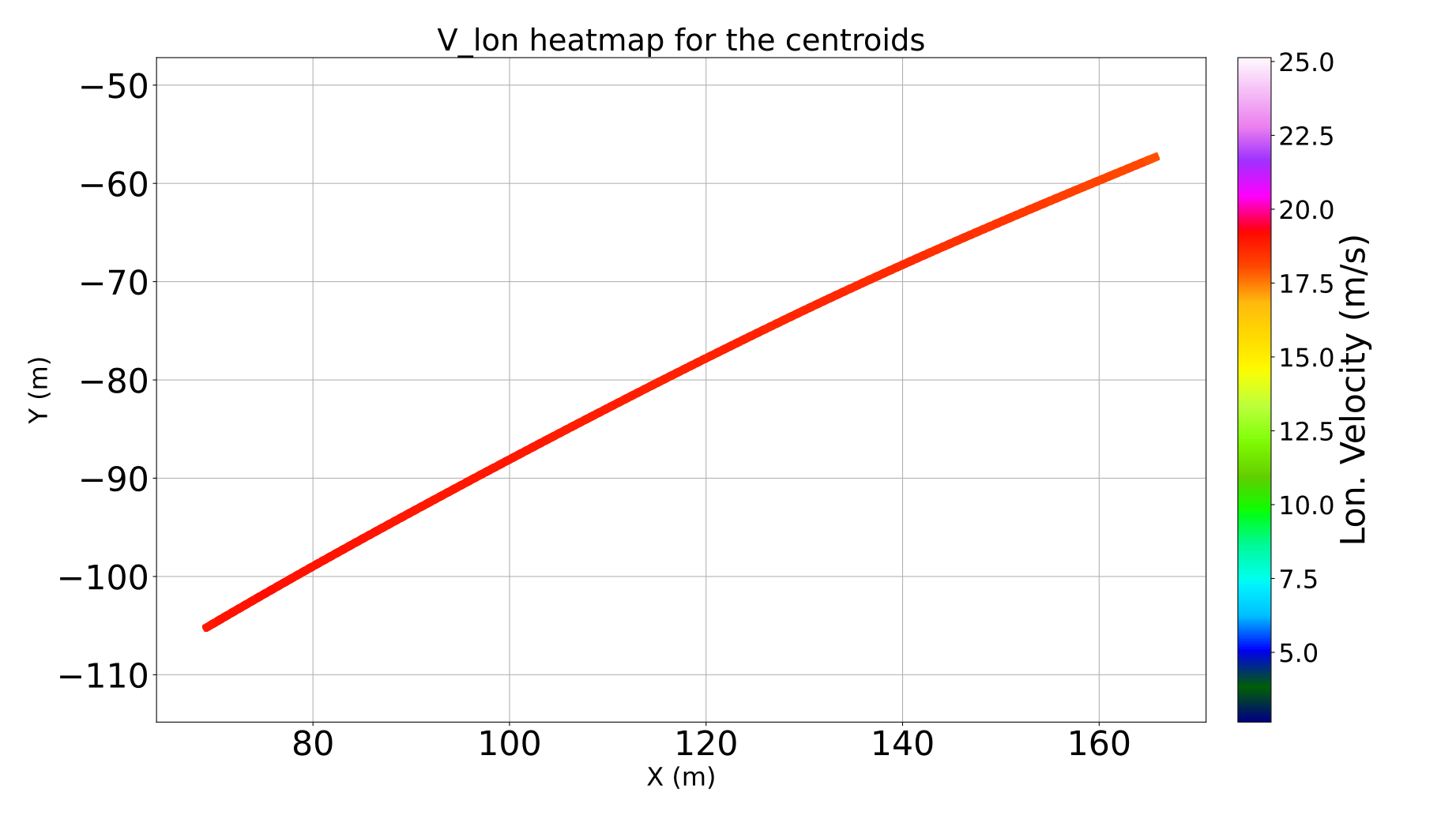} 
			}
		\end{subfigure}
		\begin{subfigure}[b]{0.59\columnwidth}
			\centering
			\adjustbox{scale=0.8, trim=1mm 0mm 4mm 1.75mm, clip}{
				\includegraphics[width=1\textwidth]{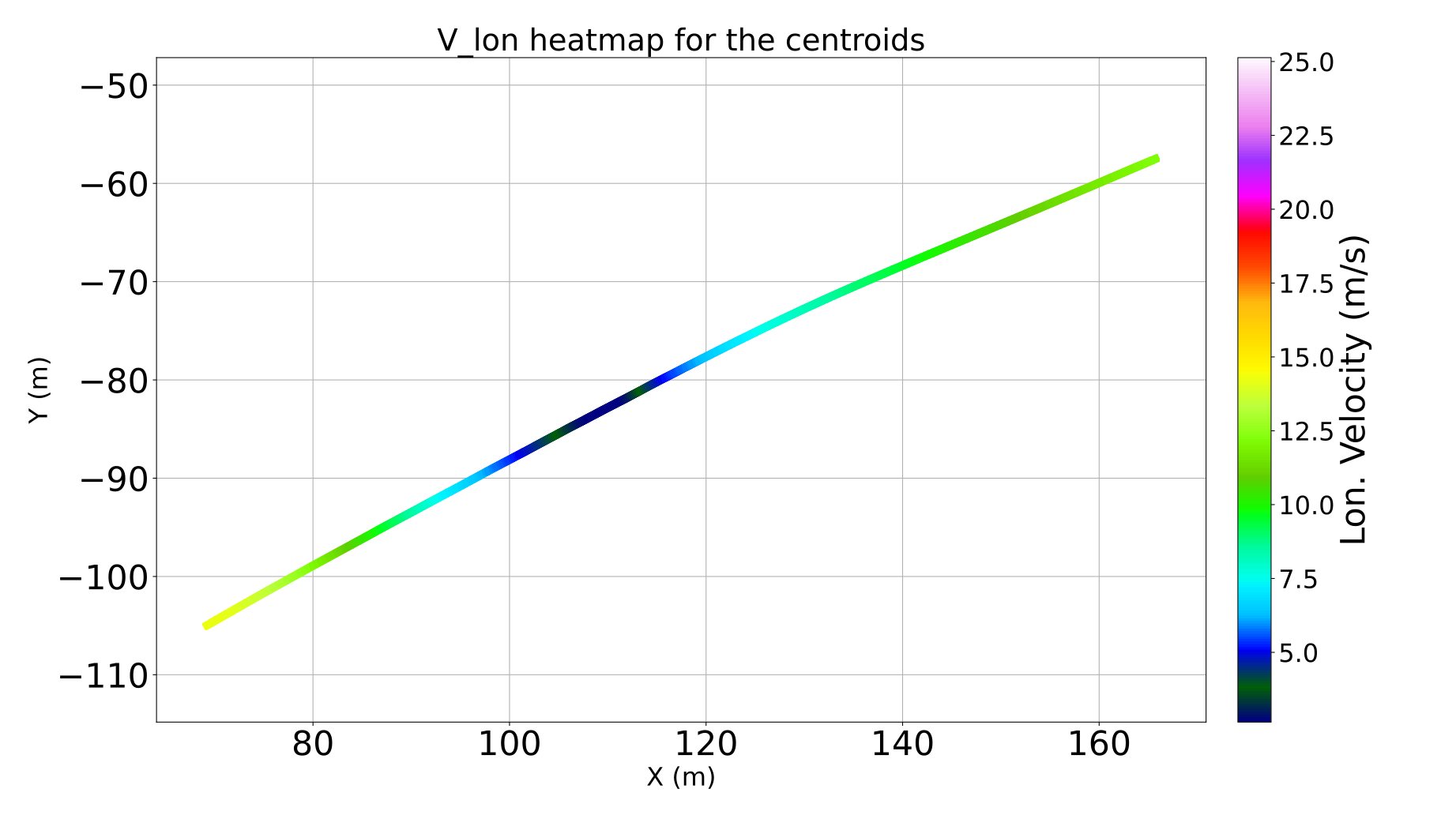} 
			}
		\end{subfigure}
		\caption{Lon. vel. clusters and centroid heat-maps from \ref{fig:7b}}
		\label{fig:10}
	\end{figure}
	
	\subsection{Related works}
	
	There is a multitude of different motivations leading to the tentative of extracting drivers' behaviors from raw data, for analysis of the discovered results or just to reproduce them. In \cite{kuderer2015learning} an inverse reinforcement learning method is used to learn driving styles from demonstration. Nine different features are considered, including jerk, acceleration, collision avoidance and lane centrality. The main goal was really to reproduce the behavior learned from multiple drivers, not to compare the differences from each driver. Another example of cluster representation approach is \cite{chai2019multipath}, where ensembles of trajectories represent modes of intent to be predicted.
	
	Initial attempts to cluster together trajectories using their velocity profiles were made by proposing different comparison metrics. In the first case, \cite{yanagisawa2006clustering} proposed to indirectly use the velocity during clustering with a Euclidean distance calculated in a window of points, and then applied it to a nearest neighbor and DTW algorithms. \cite{kerper2011driving} actually cluster the data from different segments of a trajectory to predict the future velocity profile and try to optimize fuel consumption; which is similar to \cite{demoura2023extraction}, with the only difference that the latter also uses the longitudinal acceleration for the clustering. However, it is known that these kind of methods are not capable of separating the profiles efficiently, as it will be shown in the results. Similar methods can also be seen in \cite{lee2007trajectory}, where a complete trajectory is separated in parts and then clustered with a DBSCAN-inspired approach. A more complete review of metric methods can be found in \cite{yuan2017review} and \cite{besse2016review}. 
	
	There are two other general approaches for driver behavior discovery: probabilistic-based and learning-based. In \cite{martinsson2018clustering}, a mixture of Hidden Markov Models (HMM) is used to represent each possible behavior profile. An Expectation-Maximization (EM) algorithm is then used to estimate the mixing coefficient and the parameters of each state for all HMMs. As a drawback this method could have wildly different number of clusters for similar scenarios, and also the meaning of each HMM's state is \textit{a priori} unknown. The work in \cite{dewei2019trajectory} follows the same approach and applies the clusters to predict behaviors at intersection. Specifically, it uses a polynomial regression mixture, where an EM algorithm calculates the allocation of each sample to a specific cluster and then it optimizes the polynomial that represent the cluster's centroid. \cite{sung2012traj} generically discover the minimal set of trajectory subsections that can represent its entirety and \cite{deo2018how} used a variational Gaussian Mixture Model (GMM) to classify the maneuvers in a highway of the surrounding vehicles with predefined class labels and motion models.
	
	A deep-learning approach to calculate the similarity of trajectories is proposed by \cite{li2018depp}, where an encoder-decoder architecture is trained to quantitatively evaluate the similarity of urban trajectories that might present noise, missing points or different sampling frequencies; however, this comparison uses a higher view-scale than the one needed in this article. In \cite{fang2021e2dtc} an end-to-end clustering approach is presented, improving the final results in comparison with the former publication. With an architecture based on the seq2seq publication \cite{sutskever2014sequence}, three different procedures are executed by the network: the trajectory tokenization in the embedding, its reconstruction by the decoder and finally the cluster classification. Following an embedded information approach than an explicit comparison between trajectories, \cite{harmening2020deep} uses two different networks to cluster the behavior of vehicles, the first one an encoder-decoder with a 3D convolution network as encoder to compact the position and time into a single representation and a second RNN-based network where the resulting hidden vector is also combined to cluster the data. And \cite{shouno2018deep} produced a topological map with nine six different parameters (lateral and longitudinal velocities included) from 59 demonstrations on simulation to optimize the t-SNE representation of the results and subsequent clustering using a GMM.
	
	All the cited methods use, in some form, the velocity into its proposed clustering, but none use the dynamic characteristics of a nonholonomic vehicle nor they separate the positional clustering influence from the inherent longitudinal profile each driver may have. Thus, the main contributions of this paper lies in the capability to cluster only the longitudinal information of vehicles to first confirm that drivers adopt a wide range of possible speed / acceleration profiles during the driving task and to (in a procedural manner) obtain such different behaviors. Section \ref{sec:2} details the similarity measure used while section \ref{sec:3} explains how the clustering is done. The results are reported in section \ref{sec:4}.
	
	\section{Dynamic analysis of longitudinal clusters}
	\label{sec:2}
	
\begin{figure}[!h]
		
		\adjustbox{scale=0.7, trim=40mm 10mm 10mm 10mm, clip, center=9.5cm}{
			\includegraphics[width=0.8\textwidth]{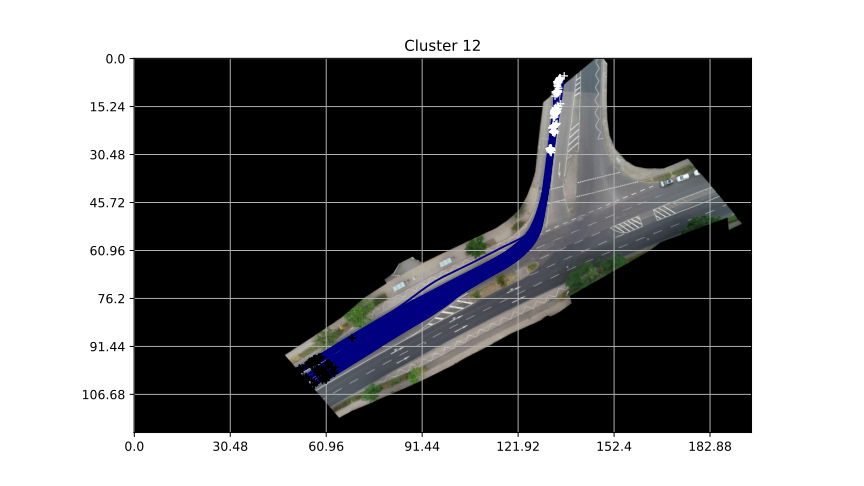} 
		}
		\caption{Example of macro-maneuver (from InD dataset \cite{bock2020})}
		\label{fig:1}
	\end{figure}

	Dealing only with the longitudinal information of trajectories assumes that the data is already structured into different macro-maneuvers\footnote{Macro-maneuver combines multiple atomic maneuvers: in Figure \ref{fig:1} the turn left can be decomposed into keep lane plus turn left. From this point on it will be referred simply as maneuvers.}. This is the scope considered here: all data used has been already processed into specific maneuvers detected in their respective scenarios. Figure \ref{fig:1} shows the maneuver 0 from scenario 0 of the InD dataset \cite{bock2020}. The clustering method to determine the set of trajectory samples for each maneuver was proposed in \cite{demoura2024hal}, which made important improvements over \cite{demoura2023extraction}. 
	
	\subsection{EKF to determine similarity}
	\label{subsec:2.1}
	
	Each trajectory considered is composed by its position ($s_{x,t}, s_{y,t}$), orientation ($\theta_t$) and the longitudinal information (velocity and acceleration, $v_{{lon}_t}, a_{{lon}_t}$), as represented in equation \eqref{eq:1}, where $\mathbb{S}_k$ indicates the sample considered and $\mathbb{T}_j$ the duration of the trajectory $j$. The state vector $\boldsymbol{x}_j(t)$ (from now onward represented as $\boldsymbol{x}_t$) and the control vector $\boldsymbol{u}_j(t)$ ($\boldsymbol{u}_t$) are defined in equations \eqref{eq:24} and \eqref{eq:25}, where $\phi_t$ represents the steering angle signal.
	
	\begin{equation}
		\label{eq:1}
		\bs{r}_j = \left[\left(s_{x,t}, s_{y,t}, \theta_t, v_{\tm{lon}, t}, a_{\tm{lon}, t}\right)_j\right], \forall t\in \mathbb{T}_j\tm{, }j\in\mathbb{S}_k
	\end{equation}
	\vspace{-5mm}
	\begin{align}
		\label{eq:24}
		\bs{x}_j(t) &= \left[s_{x,t}, s_{y,t}, \theta_t, v_{\tm{lon}, t}\right]\\
		\label{eq:25}
		\bs{u}_j(t) &= \left[a_{\tm{lon}, t}, \phi_t\right]
	\end{align}
	
	KF-derived methods are used for prediction under the assumption that observation and measurement noises can be approximated by Gaussians \cite{thrun2005probabilistic}, enabling the membership calculation from a trajectory to a certain cluster. Smoothing is also possible, since the entire trajectory is known, but unnecessary given how the EKF will be used and also the higher computational demand in comparison. Equations \eqref{eq:2}, \eqref{eq:3} and \eqref{eq:6}  through \eqref{eq:8} show the EKF framework, with $\bs{z}_t$ being the observation vector. Convergence is assured according to \cite{krener2002convergence}.
	
	\begin{align}
		\label{eq:2}
		\bar{\bs{\mu}}_{t+1} &= \mathcal{T}(\bs{\mu}_{t},\bs{u}_t)\\
		\label{eq:3}
		\bar{\bs{\Sigma}}_{t+1} &= \bs{G}_{t+1}\bs{\Sigma}_{t}\bs{G}_{t+1}^{T}+\bs{R}
	\end{align}
	\vspace{-5mm}
	\begin{align}
		\label{eq:6}
		\bs{K}_{t+1} &= \bar{\bs{\Sigma}}_{t+1}\bs{H}_{t+1}^T\cdot(\bs{H}_{t+1}\bar{\bs{\Sigma}}_{t+1} \bs{H}_{t+1}^T+\bs{Q})^{-1}\\
		\label{eq:7}
		\bs{\mu}_{t+1}  &= \bar{\bs{\mu}}_{t+1}  + \bs{K}_{t+1}(\bs{z}_{t+1}-\bs{h}(\bar{\bs{\mu}}_{t+1} ))\\
		\label{eq:8}
		\bs{\Sigma}_{t+1}  &= (\bs{I}-\bs{K}_{t+1}\bs{H}_{t+1})\bar{\bs{\Sigma}}_{t+1} 
	\end{align}
	
	The prediction model for the task at hand, $\mathcal{T}(\bs{\mu}_{t},\bs{u}_t)$, can be seen in \eqref{eq:4}: the kinematic bicycle model without slippage for vehicle nonholonomic movement using front-wheel driving, where $l_j$ is the wheelbase of the vehicle. The matrix $G$ in equation \eqref{eq:3} is the Jacobian from equation \eqref{eq:4} and can be seen in \cite{demoura2021governing} (H as well).
	
	\begin{align}
		\label{eq:4}
		\begin{cases}
			\dot{x}_{t+1}		&=	v_{t}\cos\theta_{t}\cos\phi_t \\
			\dot{y}_{t+1}		&=	v_{t}\sin\theta_{t}\cos\phi_t\\
			\dot{\theta}_{t+1}	&=	\frac{v_{t}}{l_{j}}\sin\phi_t\\
			\dot{v}_{t+1}		&=	a_{\tm{lon,t}}
		\end{cases}
	\end{align}
	
	The comparison is achieved using the trajectory from the sample being considered, $\bs{x}_t$, to calculate the steering angle $\phi_t$ and the acceleration used in $\bs{u}_t$ at equation \ref{eq:2} for the prediction step from the current cluster centroid. Multiple steps are predicted until the update, where the real centroid trajectory serves as observation.
	
	Equations \eqref{eq:7} and \eqref{eq:8} are the update correction for the prediction. Since the trajectory data used here was acquired with a 25Hz system (drone camera), each data time-step takes 40ms; the EKF prediction is executed in 10ms steps (four predictions between data-points available) and it is updated every 80ms. This design choice is important to propagate the errors from a sample that should not be in the cluster being tested. 
	
	To calculate the steering angle, the controller in \eqref{eq:5} was implemented from \cite{hoffmann2007autonomous}. $k$ is the gain, $d_{\tm{lat}}$ is the lateral distance between the current position and the target point, which is the next iteration point and $\theta_{\tm{tr}}$ the reference direction (calculate using the current and next point of the trajectory). The controller results in a global asymptotically stable equilibrium at $d_{\tm{lat}} = 0$ for $v_{\tm{lon,t}} > 0$ and $0 < (\theta_{\tm{tr}}-\theta_{t}) < \frac{\pi}{2}$ \cite{hoffmann2007autonomous}.
	
	\begin{equation}
		\label{eq:5}
		\phi_{t+1} = (\theta_{\tm{tr}}-\theta_{t}) + \arctan\left(\frac{k\cdot d_{\tm{lat}}}{v_{\tm{lon,t}}}\right)
	\end{equation}
	
	\subsection{Membership calculation}
	\label{subsec:2.2}
	
	The Mahalanobis distance $d_{M,t}$, equation \eqref{eq:9} is calculated after a certain number of sequential predictions, creating a gap between the real centroid state and the one predicted with the sample information. As it happens, this distance is distributed according to the $\chi^2_n(\delta^2)$ distribution \cite{hashemi2019generalized}, specifically a centered chi-squared with four degrees of liberty: $\chi^2_4$. A very practical use of this distance can be seen in \cite{rodriguez2010visual} where it is used to verify the integrity of a region of interest to detect moving objects using a LIDAR, or in \cite{chang2014robust}, using the distance to increase the robustness of the estimations done by an EKF implementation.
	
	\begin{equation}
		\label{eq:9}
		d_{M,t}^{2}=(\bs{z}_t-\bs{\mu}_t)^T\cdot \bs{\Sigma}_{t}^{-1}\cdot(\bs{z}_t-\bs{\mu}_t) \sim \bs{\chi}^2_n
	\end{equation}
	
	The membership of a certain sample trajectory can then be calculated before the update step using the distance given by \eqref{eq:9}, with $z_t$ being the state of the centroid trajectory at the current time-step. At time $t$, the membership probability from the sample to the cluster represented by the centroid $\bs{x}_{c_i}$ is given by equation \eqref{eq:10}. The probability at the right side of \eqref{eq:10} is the cumulative distribution function (c.d.f.) from $\chi^2_4$, representing the probability that the distance between the estimated state and the observed one is higher than what was obtained by the EKF.
	
	\begin{equation}
		\label{eq:10}
		p(\bs{x}_t\in\mathcal{C}_i | \bs{x}_{c_i}) = 1 - p(d \geq d_{M,t}^{2})
	\end{equation}
	
\begin{figure}[h]
		\centering
		\adjustbox{scale=0.8, trim=2mm 0mm 0mm 4.mm, clip}{
			\includegraphics[width=0.5\textwidth]{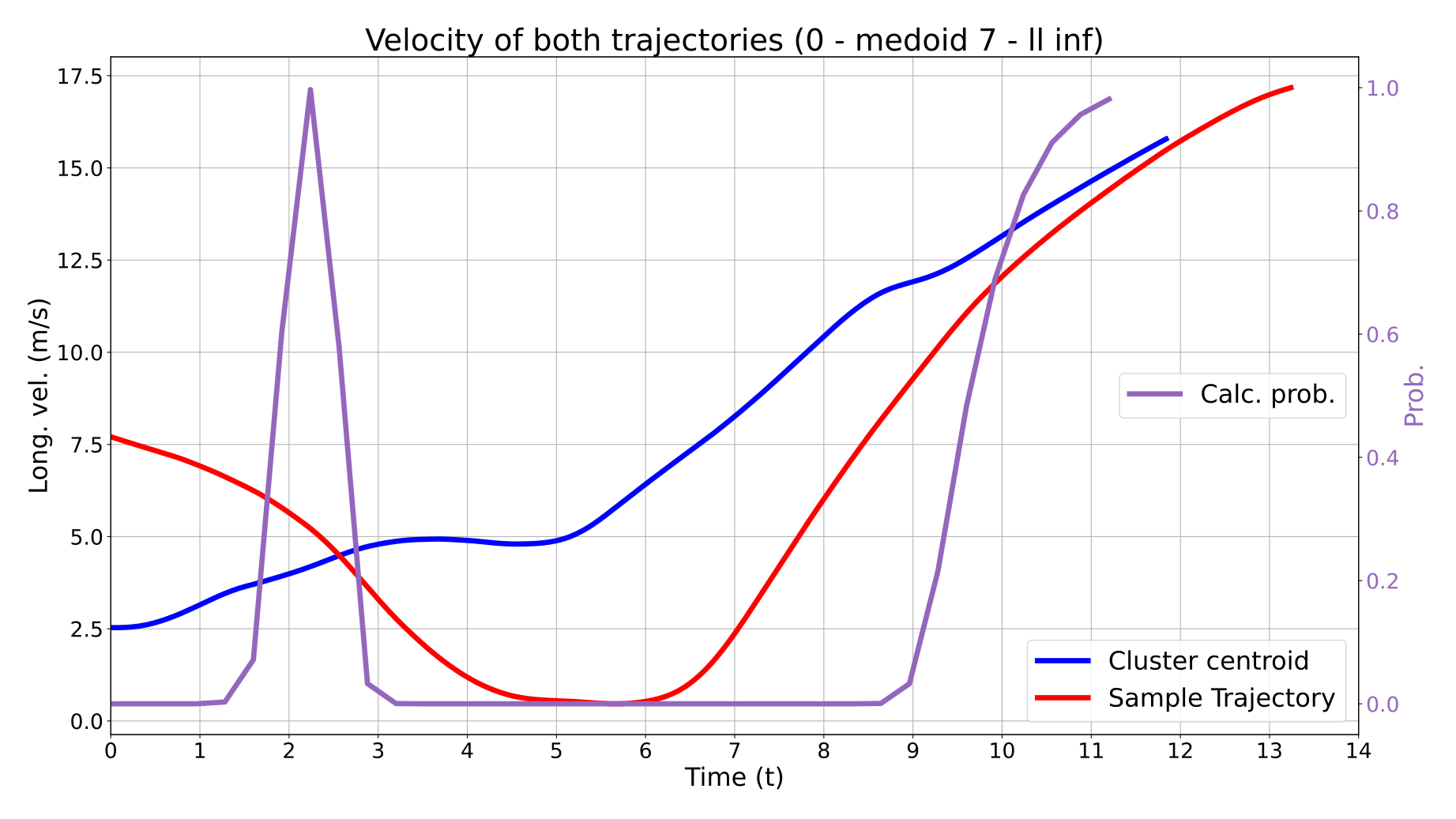} 
		}
		\caption{Two trajectories from different clusters}
		\label{fig:2}
	\end{figure}

\begin{figure}[h]
		\centering
		\adjustbox{scale=0.8, trim=2mm 0mm 0mm 5mm, clip}{
			\includegraphics[width=0.5\textwidth]{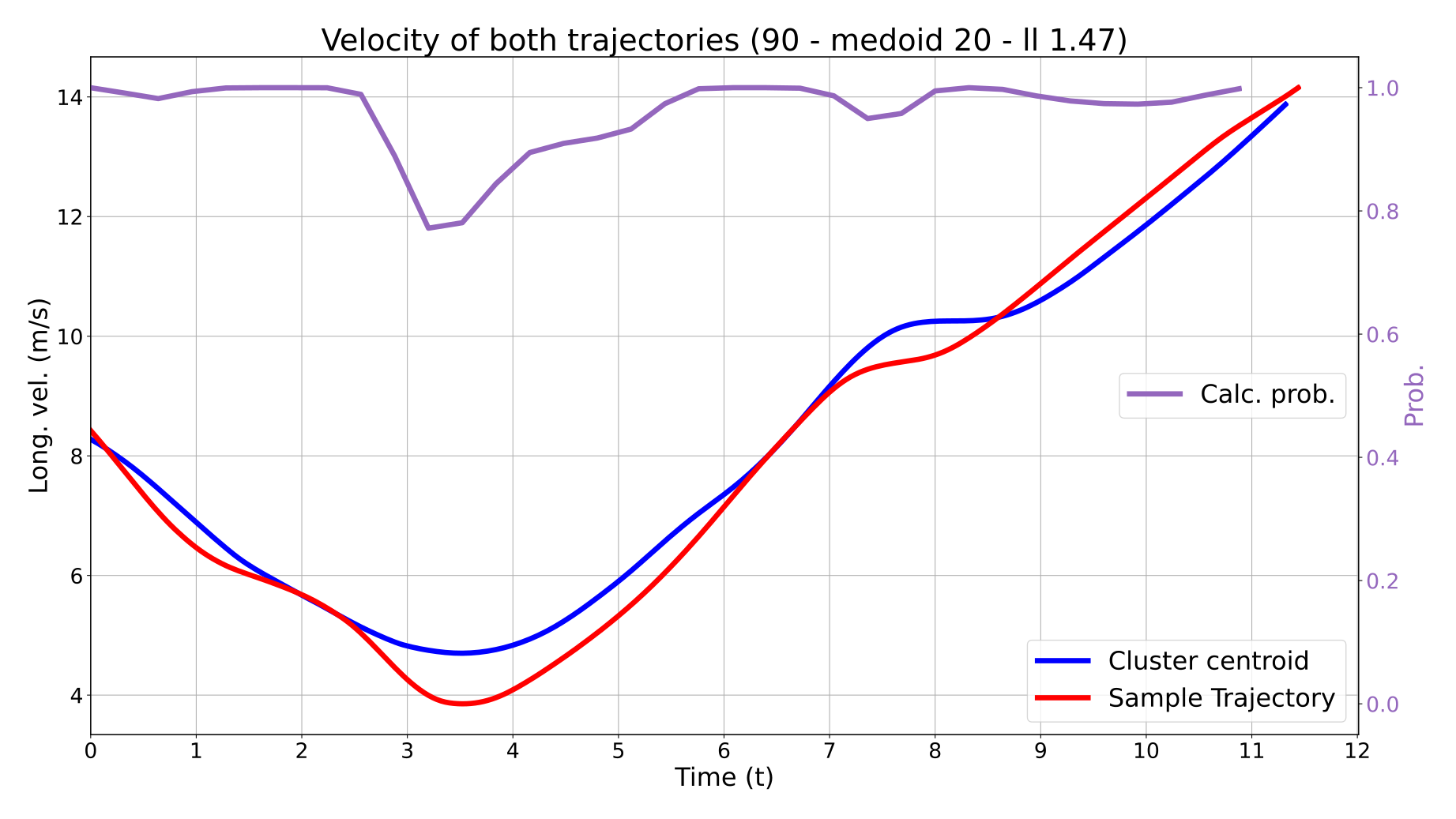} 
		}
		\caption{Two trajectories from the same cluster}
		\label{fig:3}
	\end{figure}

	As the final product of the comparison, $p(\bs{x}_t\in\mathcal{C}_i | \bs{x}_{c_i})$ will look something like Figure \ref{fig:2} for the case where the trajectory tested (in red) does not belong to the cluster whose centroid has a longitudinal velocity profile shown in blue. On the opposite, Figure \ref{fig:3} shows a sample trajectory that does belong to the cluster, fact made clear by the high probability calculated and displayed in purple.
	
	\section{Optimizing cluster distribution}
	\label{sec:3}
	
	Two different quantities will be evaluated by the proposed algorithm: the number of clusters in the studied data and the similarity between each element inside the cluster, which is represented by $p(\bs{x}_t\in\mathcal{C}_i | \bs{x}_{c_i})\tm{ , for t} \in \mathbb{T}_j$, the probability that at time $t$ the sample is similar to the cluster's centroid. This membership probability is not a practical metric to work with the expectation-maximization (EM); the negative log-likelihood (equation \eqref{eq:12}), with $\bs{x}_{c_i}$ being the centroid of $\mathcal{C}_i$ ($c_i$ indicating the element from $\mathcal{C}_i$ that is the centroid), compresses the time series into one scalar.
	
	\begin{equation}
		\label{eq:12}
		nll(\bs{x}_j|\bs{x}_{c_i}) = \sum_{t=0}^T -\ln\left[p(\bs{x}_t\in\mathcal{C}_i | \bs{x}_{c_i})\right]
	\end{equation}
	
	Another advantage from the equation \eqref{eq:12} form is that if the probability at any point in time goes to zero, \textit{nll} goes to infinity, thus being a clear marker that the sampled trajectory evaluated does not belong in the cluster being tested.
	
	\subsection{Negative log-likelihood minimization}
	\label{subsec:3.1}
	
	Following the structure of an EM method, the first step is to minimize $nll(\bs{x}_j|\bs{x}_{c_i})$ for each sample being considered using all the current centroids. Simply put, to search for the best cluster possible for each sample. Considering $p(\bs{x}_j \in \mathcal{C}_i | \bs{x}_{c_k})$ as the probability that the element $\bs{x}_j$ belongs to the cluster $\mathcal{C}_i$ given $\bs{x}_{c_k}$ as centroid, the EM algorithm consists in the expectation phase, equation \eqref{eq:13} and the maximization step in equation \ref{eq:14}. Firstly, the expectation of $\bs{x}_i$ belonging to cluster $\mathcal{C}_i$ given that the centroid used was $\bs{x}_k$ is calculated; and finally, with the cluster elements defined the calculated expectation expression is maximized by choosing the optimal centroid $\bs{x}_k$ for each cluster \cite{mclachlan2007algorithm}.
	
	\begin{align}
		\label{eq:13}
		\mu_{ll} &= \mathbb{E}_{\mathcal{C}_i|\bs{x}_j,\bs{x}_{c_k}}\left[\ln p(\bs{x}_j \in \mathcal{C}_i | \bs{x}_{c_k})\right]\\
		\label{eq:14}
		\bs{x}_{c_k} &= \argmax_{\bs{x}_{c_k}}\left[ \mathbb{E}_{\mathcal{C}_i|\bs{x}_j,\bs{x}_{c_k}}\ln p(\bs{x}_j \in \mathcal{C}_i | \bs{x}_{c_k})\right]
	\end{align}
	
	Since the EKF comparison presented in the subsection \ref{subsec:2.2} calculates $p(x_j \in \mathcal{C}_i | x_{\mathcal{C}_i})$, this operation becomes a hard clustering (in opposition to a soft clustering like C-means, where the samples are assigned multiple partial membership probabilities instead of a single one). Hence, equation \eqref{eq:15} displays the allocation of elements in the best cluster possible by minimizing the negative log-likelihood and equation \eqref{eq:16} shows the selection of the new centroid by minimizing the sum of log-likelihoods from the other elements of the cluster.
	
	\begin{align}
		\label{eq:15}
		\mathcal{C}_k(\bs{x}_j) &= \argmin_{\mathcal{C}_k \in \mathcal{C}} \left[-\ln p(\bs{x}_j \in \mathcal{C}_k | \bs{x}_{c_k})\right]\tm{, } \forall \bs{x}_j\in \mathcal{C}_k\\
		\label{eq:16}
		x_{c_k} &= \argmin_{x_{\mathcal{C}_k}}\left[ \sum_{x_k}-\ln p_t(x_i, \mathcal{C}_k | x_{c_k})\right]
	\end{align}
	
	From now onward the probability $p(\bs{x}_j \in \mathcal{C}_k | \bs{x}_{c_k})$ will be expressed by $p(\bs{x}_j | \bs{x}_{c_k})$. 
	
	\subsection{Splitting and merging clusters}
	\label{subsec:3.2}
	
	Until this point, only half of the problem was covered, hence the uncertainty about the number of clusters needs to be addressed. At the beginning of the allocation-minimization process proposed in subsection \ref{subsec:3.1}, and every time that a cluster member or more have an \textit{nll} equal to infinity, it is reallocated to the next best cluster. If all possible allocations result in infinity, then another cluster is created just for this element. This is the first mechanism to change the number of clusters. 
	
	In addition, two other criteria are necessary, one to split the clusters, and another to merge them. The method that will be proposed came from the realization that one could use the Kullback-Liebler (KL) divergence (equation \eqref{eq:17}) to compare the similarity of cluster(s) before and after splitting a cluster (or merging two), as \cite{saraiva2019data} does. In \eqref{eq:17} both $P$ and $Q$ are probability distributions defined on the same sample space.
	
	\begin{equation}
		\label{eq:17}
		D_{\tm{KL}}(P||Q) = \sum_{x\in \mathcal{X}}P(x)\cdot \ln\left(\frac{P(x)}{Q(x)}\right)
	\end{equation}
	
	To test if a cluster should be split in two, the elements are separated using a simplified k-means with k equal to $2$, establishing $\mathcal{C}_{k}^{[0]}$ and $\mathcal{C}_{k}^{[1]}$, equation \eqref{eq:20}. New centroids are chosen ($\mathcal{C}_{k}^{[0]'}$ and $\mathcal{C}_{k}^{[1]'}$) for both sets and the membership probabilities are calculated ($Q^{[0]}$ and $Q^{[1]}$). The KL divergence is applied on the distributions $\mathcal{C}_{k}^{[i]}$ and $\mathcal{C}_{k}^{[i]'}$ in equation \eqref{eq:18}, considering that the membership probabilities are actually dependent on time, as indicated by the term $\bs{x}_t$ in the probability $p(\bs{x}_t|\bs{x}_{\mathcal{C}_{k}^{[i]'}})$.
	
	\begin{align}
		\label{eq:20}
		\mathcal{C}_{k} &= \mathcal{C}_{k}^{[0]} + \mathcal{C}_{k}^{[1]}\\
	\end{align}
	\vspace{-5mm}
	\begin{multline}
		\label{eq:18}
		D_{\tm{KL}}^t(\mathcal{C}_{k}^{[i]}||\mathcal{C}_{k}^{[i]'}) =\\ \sum_{\bs{x}_j\in \mathcal{C}_{k}^{[i]}}p\left(\bs{x}_t|\bs{x}_{c_{k}^{[i]}}\right)\cdot 
		\ln\left(\frac{p\left(\bs{x}_t|\bs{x}_{c_{k}^{[i]}}\right)}{p(\bs{x}_t|\bs{x}_{c_{k}^{[i]'}})}\right)
	\end{multline}  
	
	Finally, if the criteria expressed by equation \eqref{eq:19} is respected, the cluster split is accepted and consolidated into the main cluster set. $\overline{D}_{\tm{KL}}$ represents the average of $D_{\tm{KL}}^t$ calculated in equation \ref{eq:18}.
	
	\begin{equation}
		\label{eq:19}
		\left| \overline{D}_{\tm{KL}}(\mathcal{C}_{k}^{[0]'}||\mathcal{C}_{k}^{[0]}) - \overline{D}_{\tm{KL}}(\mathcal{C}_{k}^{[1]'}||\mathcal{C}_{k}^{[1]})\right| \geq t_{\tm{KL}}
	\end{equation}  
	
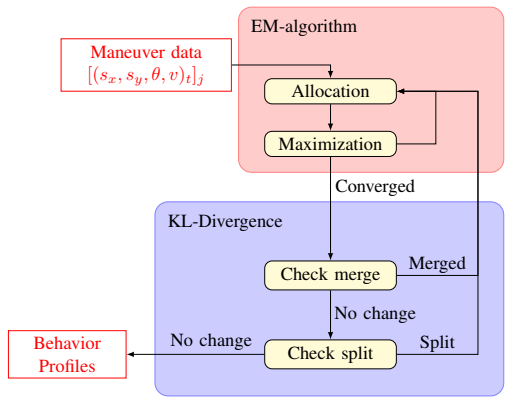
\begin{figure}[!h]
		\begin{center}
			\adjustbox{scale=0.7, trim=0mm 0mm 0mm 0mm, clip}{
				\begin{tikzpicture}
					
					\tikzstyle{inout}	= [rectangle,draw=red,fill=white!50,text=red, minimum width=2cm, text width=3cm, align=center]
					\tikzstyle{back}	= [rounded corners=0.25cm]
					\tikzstyle{box}		= [rectangle, draw=black, fill=yellow!20, text=black, minimum width=2.5cm, rounded corners]
					\tikzstyle{main}	= [rectangle, draw=red, fill=yellow!20, text=red, minimum width=2.5cm, rounded corners]
					\tikzstyle{dummy}	= [circle, radius=0cm]
					
					\tikzstyle{down}	= [->,>=latex]
					
					\node[inout]	(envi)	at (-3.5,6.5)	{Maneuver data\\ $[(s_x,s_y,\theta,v)_t]_j$};
					
					\node[]			(em) 	at (-0.5,7.2)	{EM-algorithm};
					\node[box] 		(alloc) at (0,6) 		{Allocation};
					\node[box]		(max)  	at (0,5)		{Maximization};
					\node[dummy]	(dum1)  at (3,4.75)		{};
					
					\node[]			(kl) 	at (-2 ,3.5)	{KL-Divergence};
					\node[box] 		(mer) 	at (0,2.5) 		{Check merge};
					\node[box]		(spl) 	at (0,1) 		{Check split};
					\node[dummy]	(dum2)  at (3,0.5)		{};
					
					\node[inout, text width=2cm] 	(vehi) 	at (-5,1)	{Behavior Profiles};
					
					\draw[down]		(envi) -- (0, 6.5) -- (alloc);
					\draw[down]		(alloc) -- (max);
					\draw[down] 	(max) -- (2,5) -- (2,6) -- (alloc);
					\draw[down] 	(max) -- (mer) node [midway, yshift=4mm, xshift=8.5mm] {Converged};
					\draw[down] 	(spl) -- (2.8,1) node [midway, yshift=2mm] {Split} -- (2.8,6) -- (alloc);
					\draw[down] 	(mer) -- (2.8,2.5) node [midway, yshift=2mm] {Merged} -- (2.8,6) -- (alloc);
					
					\draw[down] 	(mer) -- (spl) node [midway, yshift=0mm, xshift=8.5mm] (midbox) {No change};
					\draw[down] 	(spl) -- (vehi) node [midway, yshift=2.5mm, xshift=3mm] (midbox) {No change};
					
					
					
					
					\begin{scope}[on background layer]
						
						\node[back, fill=red!20, draw=red!50, fit=(em) (alloc) (max) (dum1)] {};
						\node[back, fill=blue!20, draw=blue!50, fit=(kl) (spl) (mer) (dum2) (midbox)] {};
						
					\end{scope}
					
				\end{tikzpicture} 
			}	
		\end{center}
		\caption{Clustering architecture proposed}
		\label{fig:11}
	\end{figure}

	The same procedure is done with the merging check-up, but with the value calculated by equation \eqref{eq:19} being smaller than the threshold $t_{\tm{KL}}$. At the end of the execution loop for equations \eqref{eq:15} and \eqref{eq:16}, all the resulting clusters are tested to check if the pairs that are most similar (\textit{nll} between their centroids) should be merged. If no changes are detected, the split hypothesis is tested (as depicted in Figure \ref{fig:11}), and if there is change, then the EM algorithm is run again. The clustering ends when there are no changes after checking for the split and the merge of all clusters.
	
	
	\section{RESULTS AND DISCUSSION}
	\label{sec:4}
	
	\subsection{Methodology}
	\label{subsec:5.1}
	
	The code used to implement and test the methods proposed was done in Python using scikit-learn and numpy. All calculations were done using a AMD 7950X processor. The noise parameters for the EKF are displayed in equations \eqref{eq:22} and \eqref{eq:23}; the trajectories were obtained from InD dataset scenario 0 (data-batches 00 to 06).
	
	\begin{align}
		\label{eq:22}
		R &= \begin{bmatrix}
			0.01	& 0.01 	& 0.005 & 0.01
		\end{bmatrix}\cdot I_4\\
		\label{eq:23}
		Q &= \begin{bmatrix}
			0.05	& 0.05 	& 0.01 & 0.1
		\end{bmatrix}\cdot I_4
	\end{align}

	\subsection{Analyzing the common clustering result}
	\label{subsec:5.2}
	
	In light of this new method to obtain vehicles' behaviors, the final results from \cite{demoura2023extraction} need to be checked. Table \ref{tab:1} shows the evaluation using the average DB score proposed in \cite{demoura2024hal} and the number of infinite \textit{nll} clusters.
	
	\begin{table}[h]
		\caption{Results using \cite{demoura2023extraction} clustering for maneuver 0} 
		\label{tab:1}
		\centering
		\scriptsize
		\setlength{\tabcolsep}{1mm}
		\adjustbox{scale=0.9}{
			\begin{tabular}{c|cc|c|cc}
				\toprule
				$n_k$ & \textbf{Avg. DB} & \textbf{nb. inf. nll.} & $n_k$ & \textbf{Avg. DB} & \textbf{nb. inf. nll.} \\
				\midrule
				2 		& 1.38 	& 74 & 16	& 0.7  	& 47	\\
				3 		& 0.92 	& 68 & 17	& 0.72 	& 33	\\
				4 		& 0.90 	& 68 & 18	& 0.68 	& 33	\\
				8 		& 0.81 	& 71 & 22	& 0.64 	& 42	\\
				9 		& 0.78 	& 72 & 23	& 0.65 	& 41	\\
				10 		& 0.83 	& 48 & 24	& 0.65 	& 41	\\
				13 		& 0.76	& 48 & 27	& 0.67 	& 39	\\
				14 		& 0.69 	& 48 & 28	& 0.60 	& 40	\\
				15 		& 0.71 	& 48 & 29	& 0.61 	& 40	\\
				\bottomrule
			\end{tabular}
		} 
	\end{table}
	
	There are 146 trajectories for this maneuver, displayed in Figure \ref{fig:1}. For any value of $n_k$ tested no less than 33 points were considered as having infinite \textit{nll}, meaning that the distance metric chosen for the clustering is not capable to capture the differences between trajectories. According to the DB score used the optimal number of clusters is 28, but this can be considered as an effect of the high cluster count, not a better clustering since the score decreases progressively with the increase of $n_k$. 
	
\begin{figure}[!h]
		\centering
		\adjustbox{scale=0.7, trim=2mm 0mm 0mm 2mm, clip}{
			\includegraphics[width=0.5\textwidth]{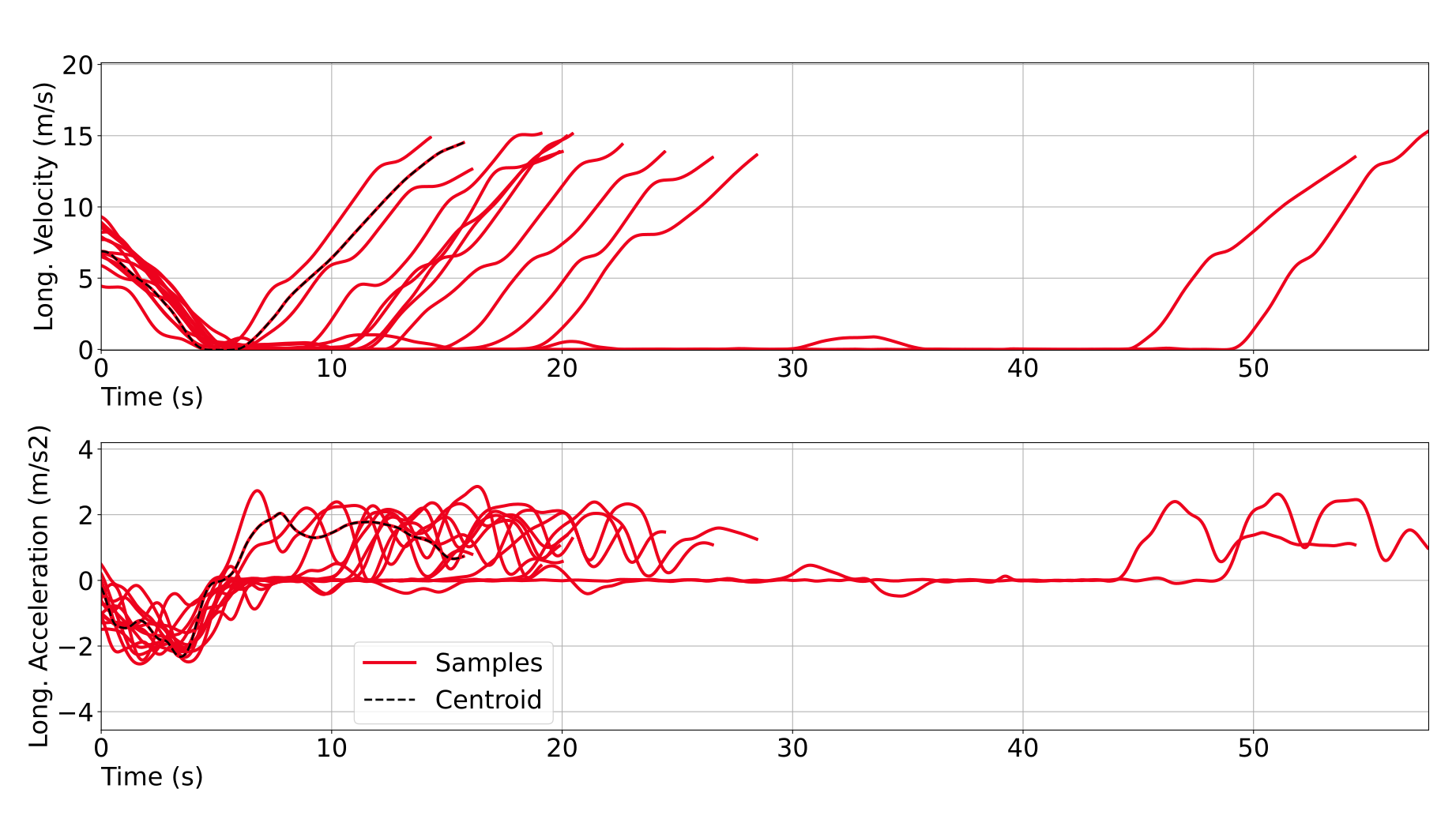} 
		}
		\caption{One of the clusters obtained with $n_k$ equal to 20 for maneuver 0 using the agglo. clustering}
		\label{fig:5}
	\end{figure}

	
	In Figure \ref{fig:5} one of the worst cases from the agglomerative clustering can be seen ($n_k=20$), and it is clear that there are multiple behaviors merged in this simple cluster. 
	
	\subsection{Full redistribution of samples}
	\label{subsec:5.3}
	
	Due to space constraints, the analysis will be restricted to only a few maneuvers, starting with the one shown in Figure \ref{fig:1}. The methods proposed in sections \ref{sec:3} and \ref{sec:4} depends upon two parameters: the initial clustering organization and the split-merge threshold, $t_{KL}$. As a starting point for the algorithm, four different clustering results were used, obtained with four methods: agglomerative clustering, partition around medoids, dissimilarity matrix clustering \cite{demoura2023extraction} and spectral clustering. The choice of $t_{KL}$ is discussed in section \ref{subsec:5.5}. All trajectories were rectified to start and to end at approximately similar points. This allows the algorithm to concentrate on the common parts of each trajectory, and avoid any detection problems from the original video source (from the InD dataset). The symbol $\mu_{ll}$ represents the average \textit{nll} from all clusters (and $\sigma_{ll}$ the standard deviation) and $n_k^f$ the final number of clusters after execution (Figure \ref{fig:11}).
	
	The best result came from the initial $n_k$ equal to 9 from the PAM, since it had the lowest $\mu_{ll}$ + $\sigma_{ll}$. It can be seen that there is still an influence from the clustering method as both the starting point and the parameter $t_{KL}$ have an importance on deciding the tightness of each cluster (lower values means more clusters). 
	
	\begin{table}[h]
		\caption{Clustering results for maneuver 0 ($n_k^i$ being the initial $n_k$ and $n_k^f$ the final one)} 
		\label{tab:3}
		\centering
		\scriptsize
		\setlength{\tabcolsep}{1mm}
		\adjustbox{scale=0.9}{
			\begin{tabular}{cccc|cccc|cccc}
				\toprule 
				$n_k^i$ & $\mu_{ll}$ & $\sigma_{ll}$ & $n_k^f$ & $n_k^i$ & $\mu_{ll}$ & $\sigma_{ll}$ & $n_k^f$
				& $n_k^i$ & $\mu_{ll}$ & $\sigma_{ll}$ & $n_k^f$\\
				\midrule
				\multicolumn{12}{c}{Agglomerative clustering}	\\
				2 & 27.32 & 38.02 & 32 & 6 & 27.01	& 41.14	& 33 & 10 & 35.11	& 49.58 & 25 \\
				3 & 27.74 & 41.84 & 33 & 7 & 31.92	& 43.64	& 27 & 11 & 35.80 	& 50.01 & 26 \\
				4 & 34.59 & 46.71 & 26 & 8 & 29.68  & 42.43	& 30 & 12 & 38.72	& 55.13 & 24 \\
				5 & 33.13 & 46.44 & 28 & 9 & 32.00  & 43.77	& 28 & -  & - 	    & - & - \\
				\midrule
				\multicolumn{12}{c}{Partition around Medoids (PAM)}	\\
				2 & 37.16 & 51.50 & 23 & 6	& 29.68	& 41.17	& 30 & 10 & 37.47 & 50.10 & 23 \\
				3 & 39.01 & 53.84 & 23 & 7	& 33.60	& 48.32	& 27 & 11 & 37.07 & 50.40 & 23 \\
				4 & 33.81 & 43.20 & 27 & 8	& 32.12	& 42.61	& 28 & 12 & 32.45 & 56.14 & 29 \\
				5 & 29.78 & 44.38 & 29 & \textbf{9}	& \textbf{27.96}	& \textbf{33.59}	& \textbf{30} & - & - 	& - & - \\
				\midrule
				\multicolumn{12}{c}{Dissimilarity matrix k-means}	\\
				2 & 31.02 & 43.52  & 29 & 6	& 31.64	& 42.46	& 28 & 10 & 26.88 & 39.07 & 31 \\
				3 & 37.84 & 48.91  & 25 & 7	& 36.51	& 49.76	& 25 & 11 & 31.60 & 43.57 & 28 \\
				4 & 36.92 & 53.99  & 24 & 8	& 39.20	& 51.69 & 23 & 12 & 27.86 & 46.26 & 32 \\
				5 & 32.32 & 42.87  & 27 & 9	& 35.56	& 50.62	& 25 & -  & -	  & -     & - \\
				\midrule
				\multicolumn{12}{c}{Spectral clustering}				\\
				2 & 38.20 & 52.11 & 23 & 6	& 29.36	& 40.66	& 31 & 10 & 29.87 & 42.07 & 30 \\
				3 & 30.35 & 43.24 & 30 & 7	& 36.29	& 50.88	& 25 & 11 & 34.04 & 48.03 & 26 \\
				4 & 30.68 & 43.37 & 29 & 8	& 34.61	& 48.27	& 26 & 12 & 34.88 & 46.51 & 25 \\
				5 & 41.56 & 50.94 & 24 & 9	& 35.67	& 45.55	& 26 & -  & -     & -     & - \\
				\bottomrule
			\end{tabular}
		} 
	\end{table}
	
\begin{figure}[!h]
		\centering
		
		\begin{subfigure}[b]{0.25\columnwidth}
			\centering
			\adjustbox{scale=3, trim=1mm 0mm 16.5mm 0mm, clip}{
				\includegraphics[width=1\textwidth]{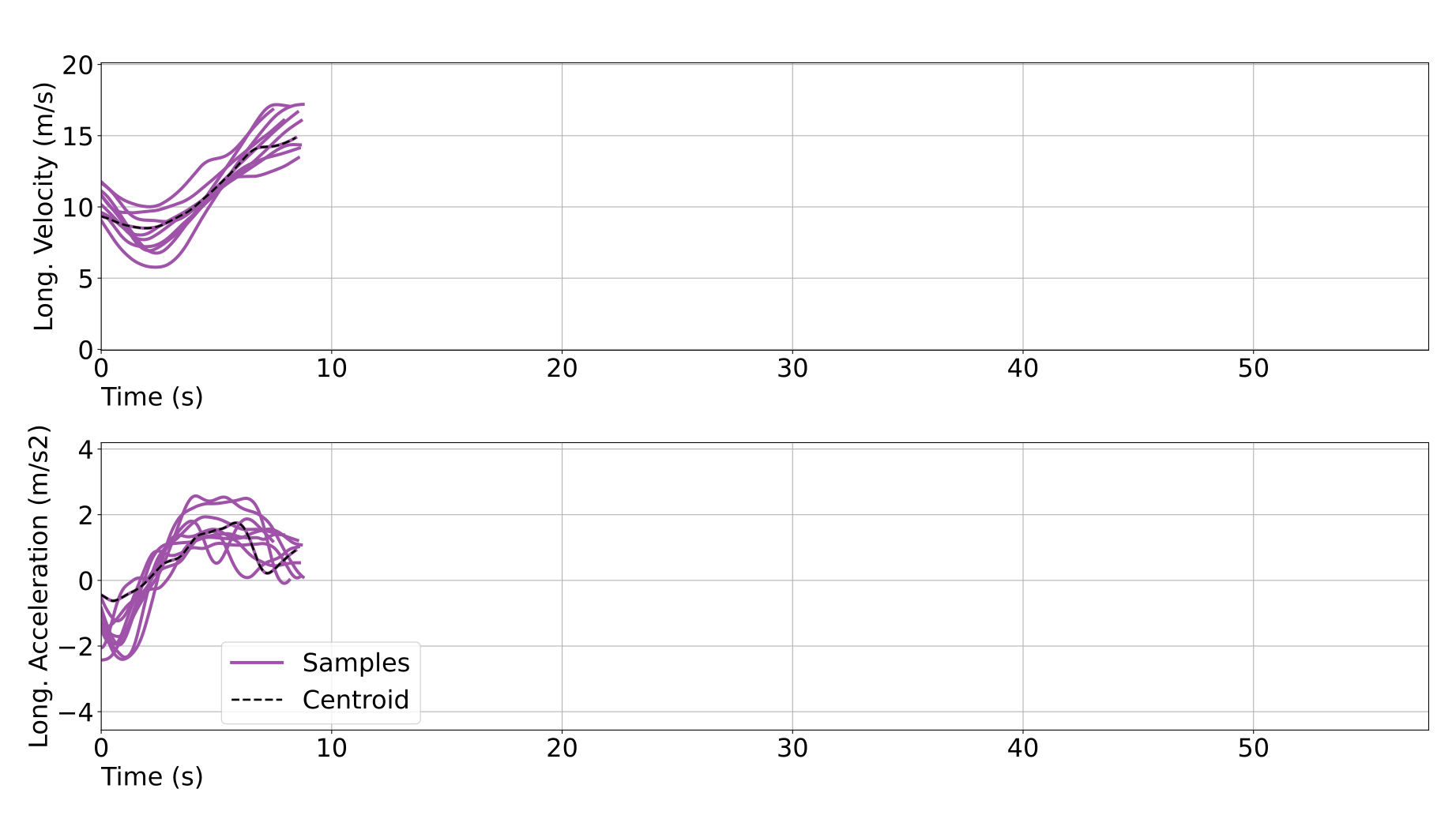} 
			}
		\end{subfigure}
		\begin{subfigure}[b]{0.3\columnwidth}
			\centering
			\adjustbox{scale=2.5, trim=2.4mm 0mm 19mm 1mm, clip}{
				\includegraphics[width=1\textwidth]{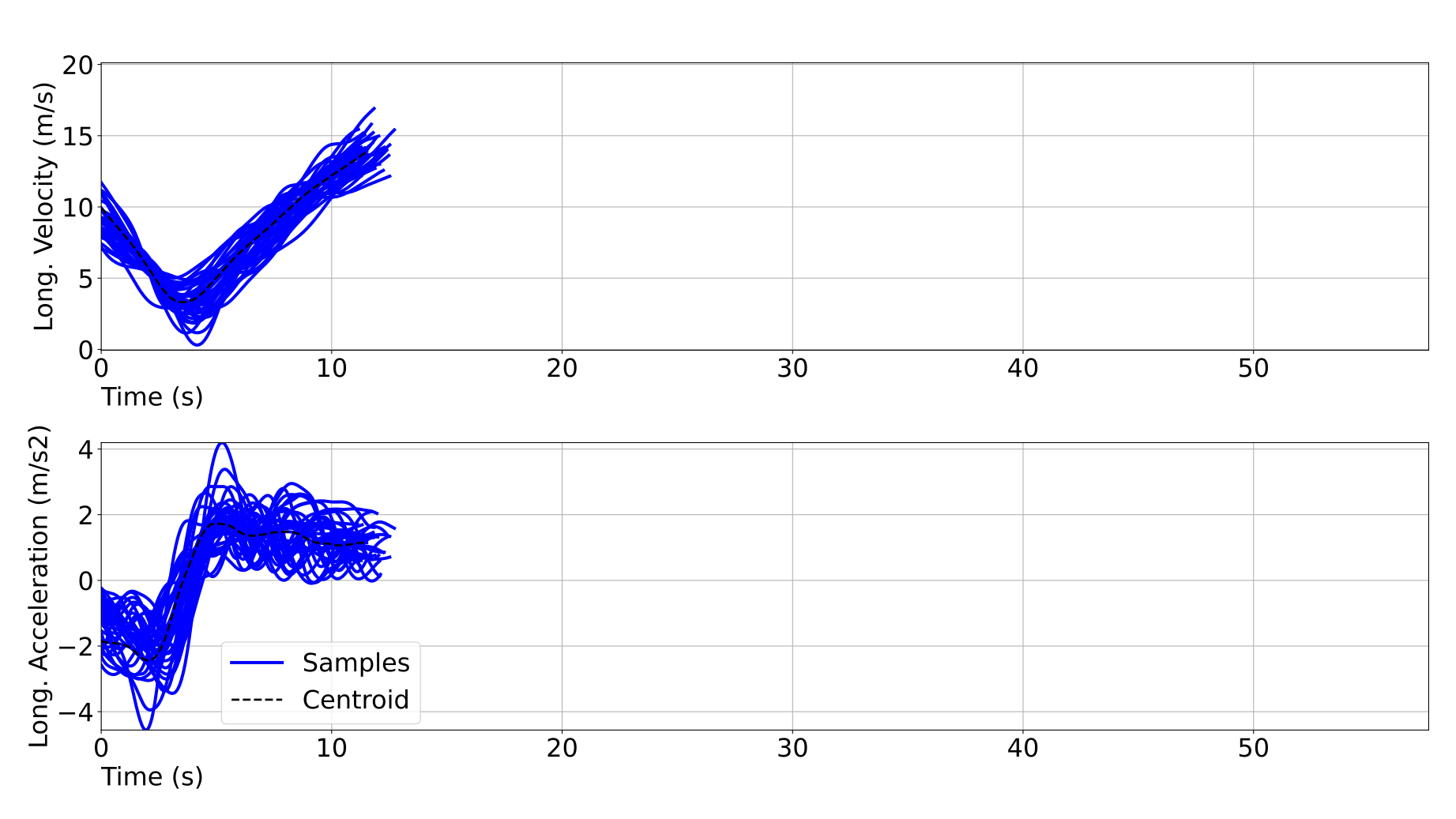} 
			}
		\end{subfigure}
		\begin{subfigure}[b]{0.38\columnwidth}
			\centering
			\adjustbox{scale=2, trim=2.5mm 0mm 17mm 1mm, clip}{
				\includegraphics[width=1\textwidth]{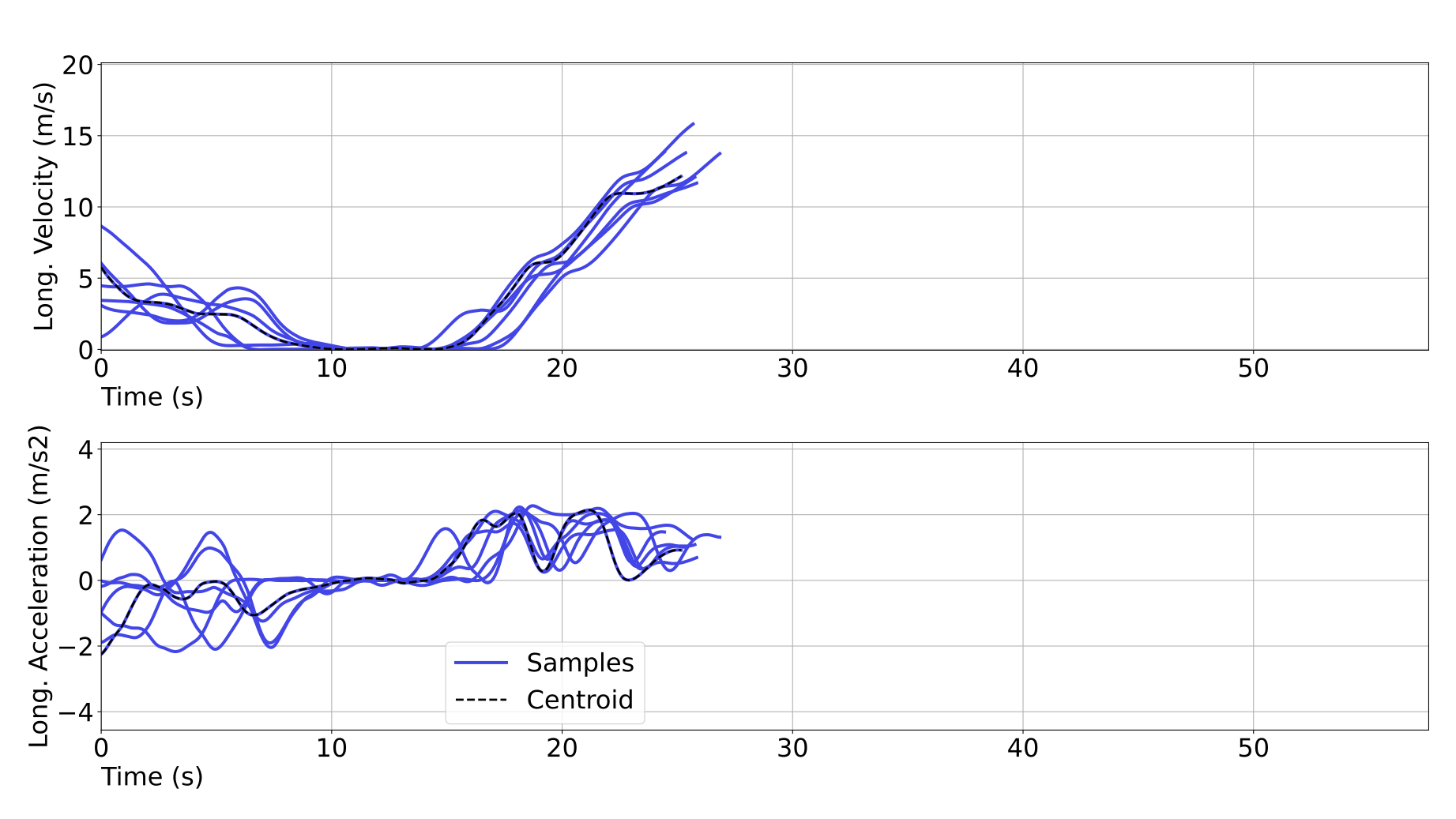} 
			}
		\end{subfigure}
		
		\caption{Four clusters from maneuver 0 result}
		\label{fig:6}
	\end{figure}

\begin{figure}[!h]
		\centering
		
		\begin{subfigure}[b]{0.32\columnwidth}
			\centering
			\adjustbox{scale=1.4, trim=1mm 0mm 10mm 1.5mm, clip}{
				\includegraphics[width=1\textwidth]{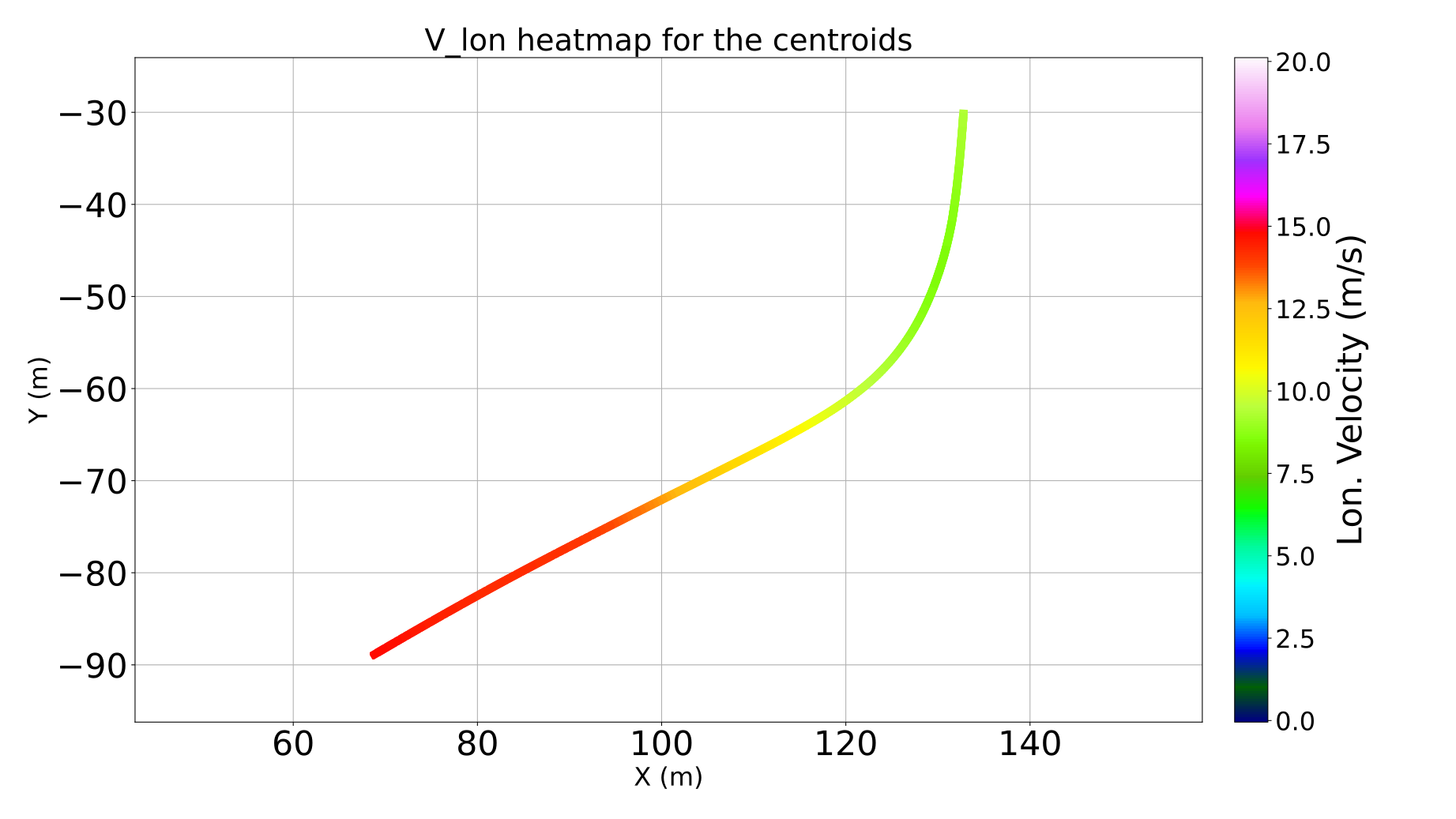} 
			}
		\end{subfigure}
		\begin{subfigure}[b]{0.32\columnwidth}
			\centering
			\adjustbox{scale=1.4, trim=7mm 0mm 10mm 1.5mm, clip}{
				\includegraphics[width=1\textwidth]{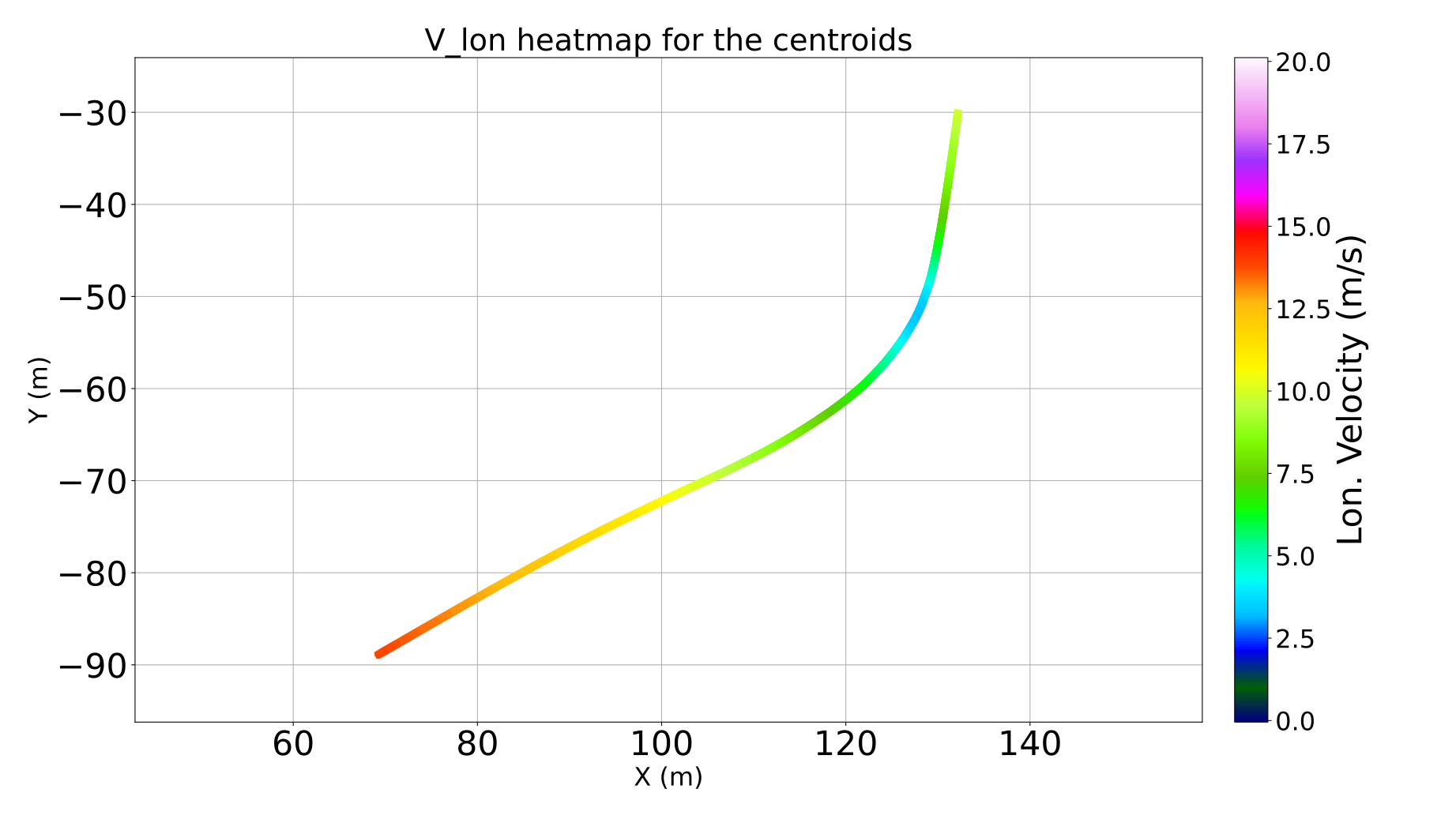} 
			}
		\end{subfigure}
		\begin{subfigure}[b]{0.33\columnwidth}
			\centering
			\adjustbox{scale=1.4, trim=7mm 0mm 4mm 1.5mm, clip}{
				\includegraphics[width=1\textwidth]{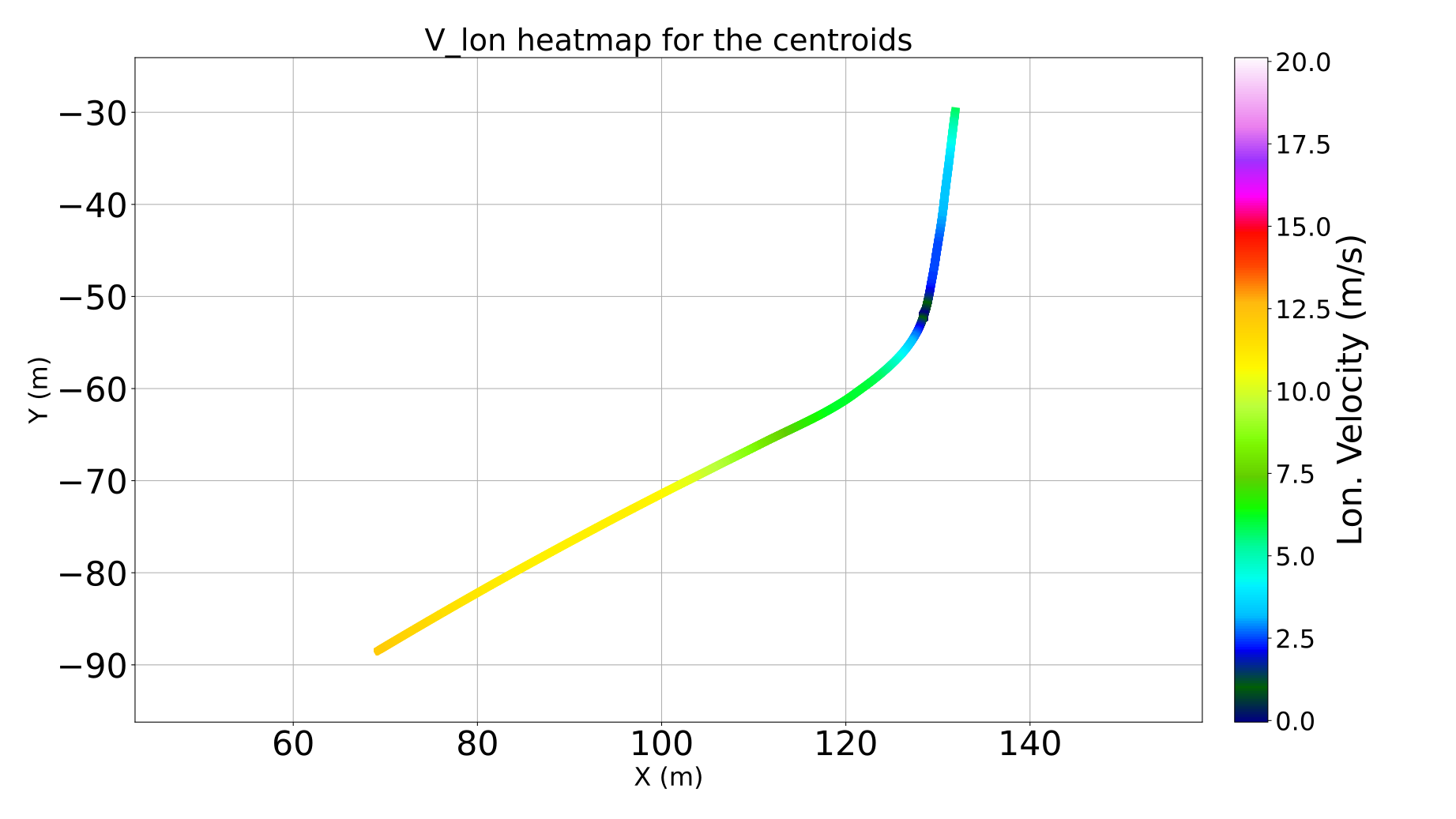} 
			}
		\end{subfigure}
		
		\caption{Lon. velocity for the centroids from Figure \ref{fig:6}}
		\label{fig:8}
	\end{figure}

	Three of the clusters are presented in Figure \ref{fig:6} together with their centroids longitudinal velocity in Figure \ref{fig:8} (please refer to the digital version for a better visualization). The differences in velocity between each cluster are clear in the Figure \ref{fig:8}, the right most plot displaying a vehicle that stopped before crossing while the leftmost plot shows a direct trajectory, without any interaction. An in-between behavior is presented by the middle plot, reduction in speed but without stopping.
	
\begin{figure}[!b]
		\begin{subfigure}[b]{0.49\columnwidth}
			\centering
			\adjustbox{scale=1.4, trim=10mm 3.5mm 0mm 3.5mm, clip}{
				\includegraphics[width=\textwidth]{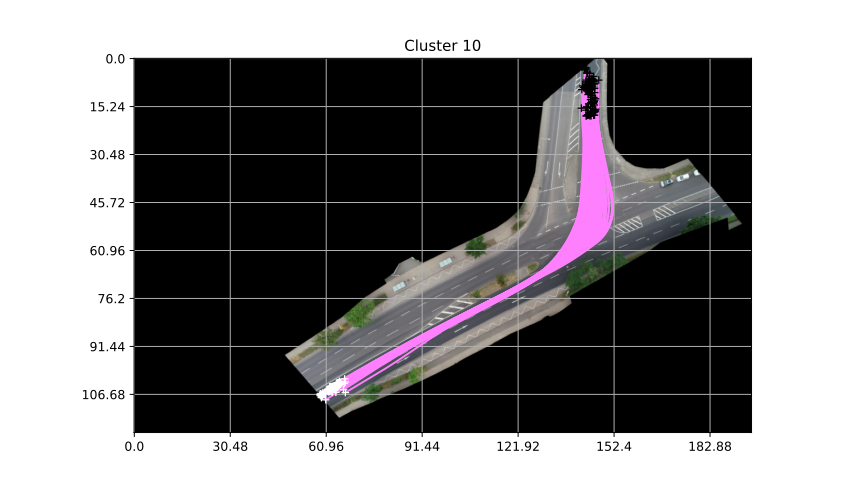} 
			}
			\caption{Turn left maneuver}
			\label{fig:7a}
		\end{subfigure}
		\begin{subfigure}[b]{0.49\columnwidth}
			\centering
			\adjustbox{scale=1.4, trim=10mm 3.5mm 0mm 3.5mm, clip}{
				\includegraphics[width=\textwidth]{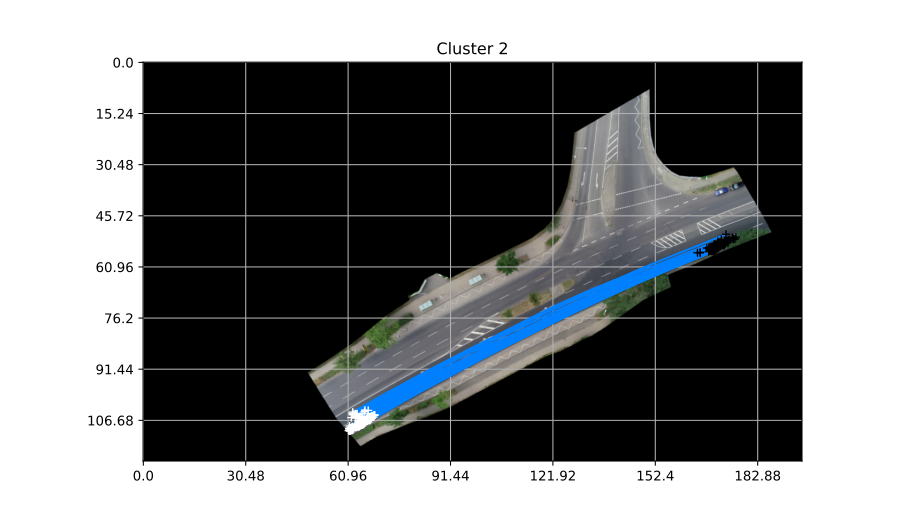} 
			}
			\caption{Straight maneuver}
			\label{fig:7b}
		\end{subfigure}
		\caption{Additional maneuvers from scenario 0}
		\label{fig:7}
	\end{figure}

\begin{figure}[!h]
		\centering
		
		\begin{subfigure}[b]{0.49\columnwidth}
			\centering
			\adjustbox{scale=1.5, trim=1mm 0mm 18mm 2mm, clip}{
				\includegraphics[width=1\textwidth]{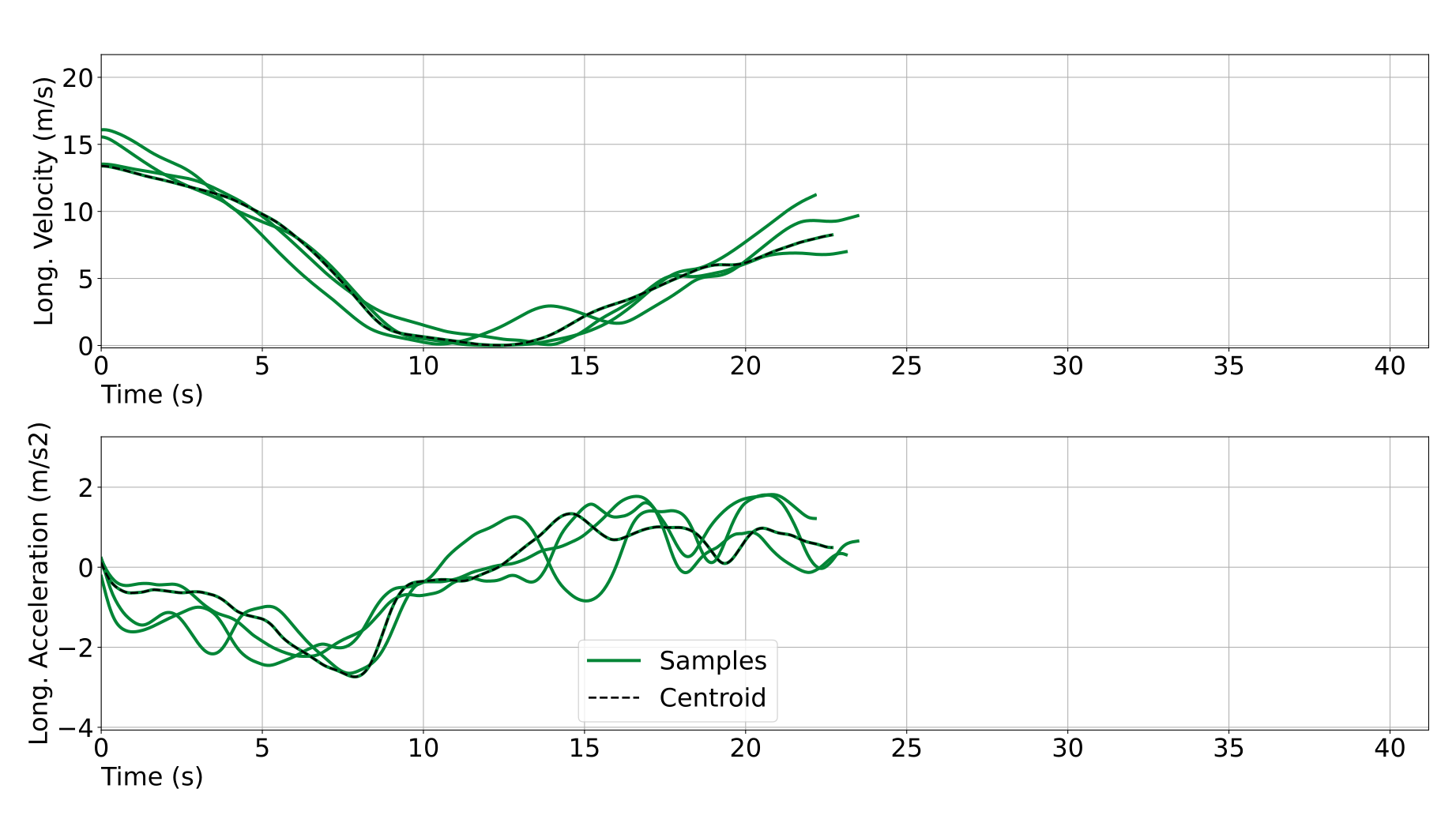} 
			}
		\end{subfigure}
		\begin{subfigure}[b]{0.49\columnwidth}
			\centering
			\adjustbox{scale=1.5, trim=1mm 0mm 22mm 2mm, clip}{
				\includegraphics[width=1\textwidth]{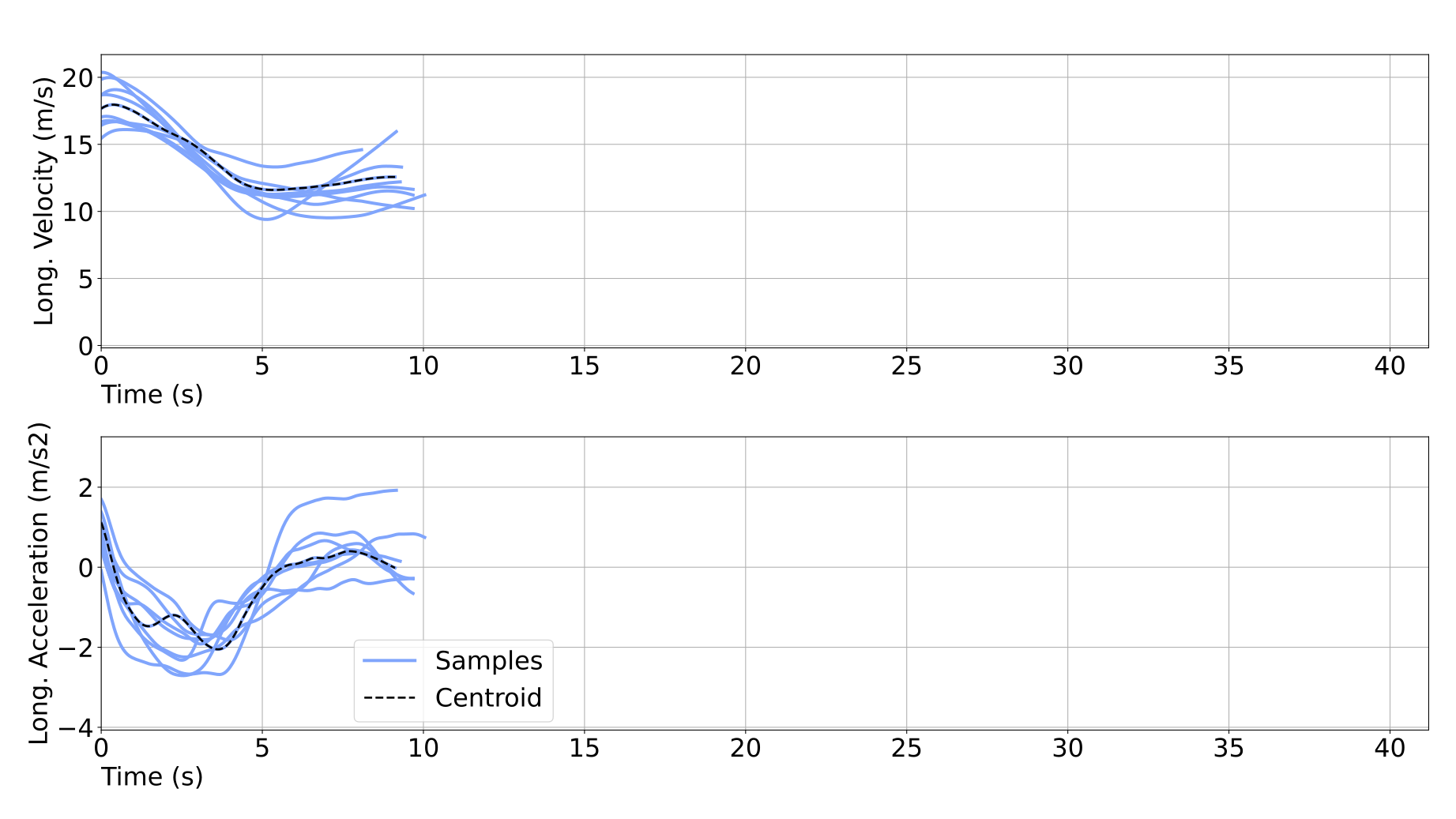} 
			}
		\end{subfigure}

		\begin{subfigure}[b]{0.49\columnwidth}
			\centering
			\adjustbox{scale=1.1, trim=1mm 0mm 14mm 2mm, clip}{
				\includegraphics[width=1\textwidth]{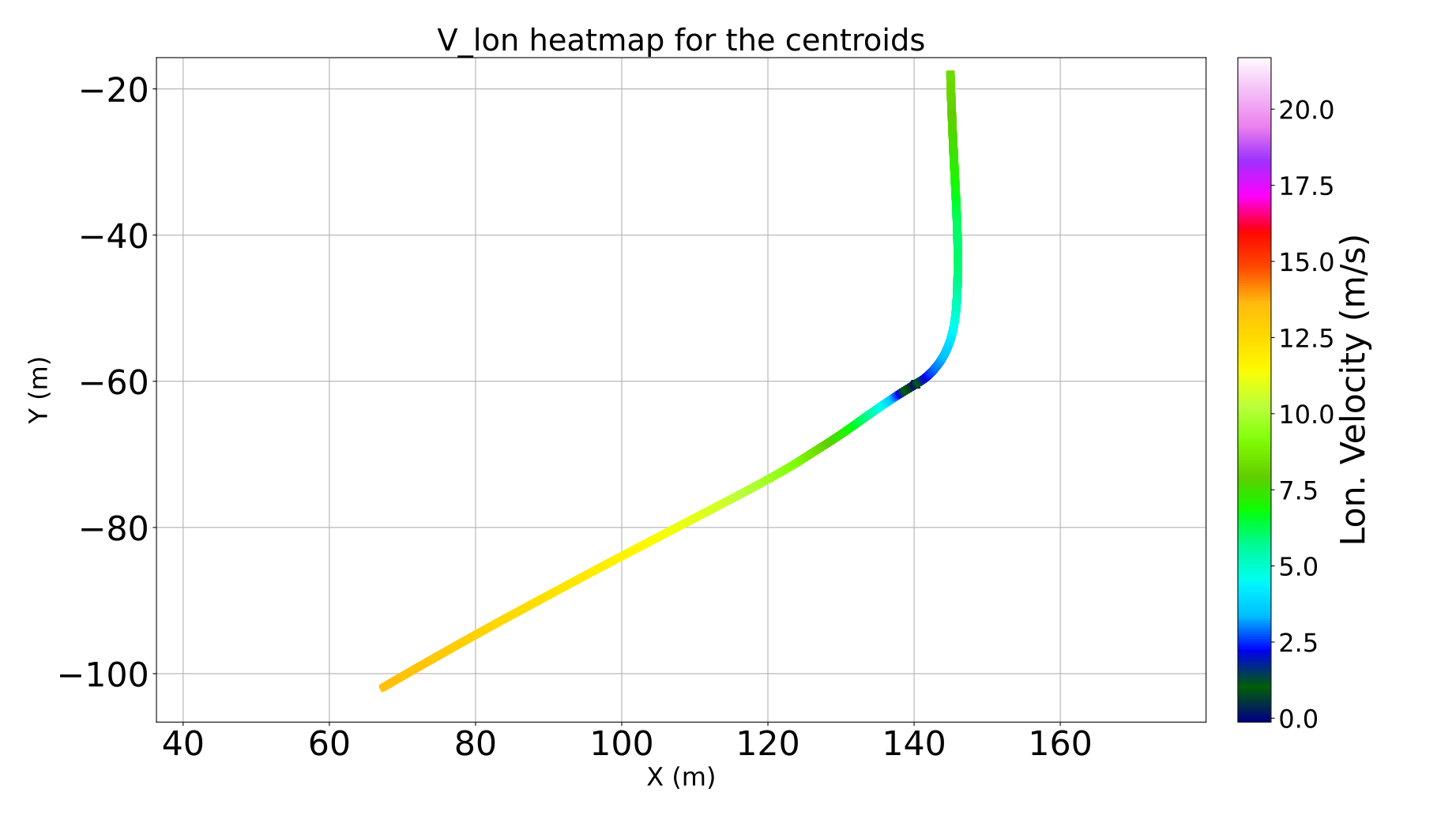} 
			}
			\caption{Stop before crossing}
			\label{fig:9a}
		\end{subfigure}
		\begin{subfigure}[b]{0.49\columnwidth}
			\centering
			\adjustbox{scale=1.1, trim=8mm 0mm 4mm 2mm, clip}{
				\includegraphics[width=1\textwidth]{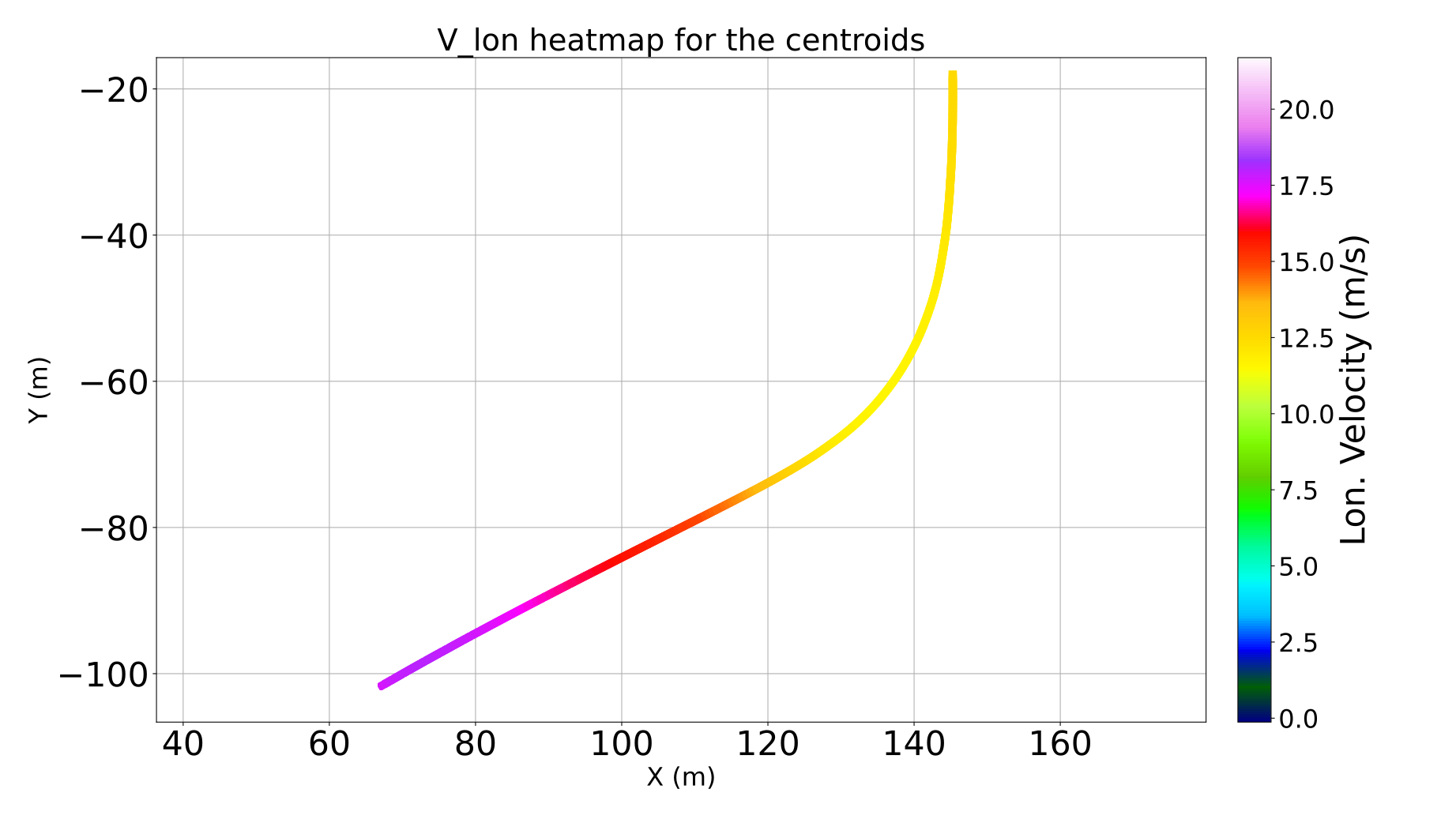} 
			}
			\caption{Crossed directly}
			\label{fig:9b}
		\end{subfigure}
		\caption{Lon. velocity profiles and heat-maps from \ref{fig:7a}}
		\label{fig:9}
	\end{figure}

	Two other maneuvers were tested: a turn left (Figure \ref{fig:7a}) and a straight maneuver (Figure \ref{fig:7b}). For the turn left the results for the agglomerating starting point are displayed in Table \ref{tab:4} and two specific clusters are shown in Figure \ref{fig:9}.
	
	\begin{table}[h]
		\caption{Results for the turn left maneuver} 
		\label{tab:4} 
		\centering
		\scriptsize
		\setlength{\tabcolsep}{1mm}
		\adjustbox{scale=0.9}{
			\begin{tabular}{cccc|cccc|cccc}
				\toprule 
				$n_k^i$ & $\mu_{ll}$ & $\sigma_{ll}$ & $n_k^f$ & $n_k^i$ & $\mu_{ll}$ & $\sigma_{ll}$ & $n_k^f$
				& $n_k^i$ & $\mu_{ll}$ & $\sigma_{ll}$ & $n_k^f$\\
				\midrule
				2 			& 22.70 		 & 36.63 		& 23	 		& 6 & 23.35 & 35.56 & 22 & 10 & 29.81 & 59.96 & 21\\
				\textbf{3} 	& \textbf{23.61} & \textbf{35.24} & \textbf{23} & 7 & 35.53 & 56.08 & 16 & 11 & 21.90 & 41.26 & 26\\
				4			& 29.19     	 & 43.63 		& 19 			& 8 & 39.26 & 76.56 & 15 & 12 & 24.88 & 43.00 & 22\\
				5 			& 27.41 		 & 42.60 		& 20 			& 9 & 39.52 & 56.61 & 14 & -  & - 	   & -     & -\\
				\midrule
			\end{tabular} 
		}
	\end{table}
	
	In Figure \ref{fig:9a} all samples stopped before crossing the intersection. Opposite to this behavior, Figure \ref{fig:9b} displays trajectories that continue only reducing their speed to execute the turn left maneuver.
	
	There is no lack of samples for the maneuver in Figure \ref{fig:7b}, containing 532 trajectories. But in this case the agglomerative starting point could not differentiate between the longitudinal characteristics, which made the job of optimize the clustering and determining the best number of clusters a lot harder. Due to this difficulty, the spectral method presented a better starting point for the clustering (Table \ref{tab:5}), given that it relies on the dimensional reduction to execute the clustering.
	
	\begin{table}[h]
		\caption{Results for the straight maneuver} 
		\label{tab:5}
		\centering
		\scriptsize
		\setlength{\tabcolsep}{1mm}
		\adjustbox{scale=0.9}{
			\begin{tabular}{cccc|cccc|cccc}
				\toprule 
				$n_k^i$ & $\mu_{ll}$ & $\sigma_{ll}$ & $n_k^f$ 
				& $n_k^i$ & $\mu_{ll}$ & $\sigma_{ll}$ & $n_k^f$
				& $n_k^i$ & $\mu_{ll}$ & $\sigma_{ll}$ & $n_k^f$\\
				\midrule
				\multicolumn{12}{c}{Agglomerative clustering}	\\
				2 		& 4.78 & 12.98 & 11 & 6 & 5.07 & 11.41 & 11 & 10 & 4.88 & 10.23 & 11\\
				3 		& 3.71 & 13.21 & 12 & 7 & 5.14 & 9.51 & 11 & 11 & 3.99 & 9.85 & 12\\
				4 		& 4.25 & 10.08 & 12 & 8 & 7.00 & 14.89 & 9 & 12 & 4.39 & 12.03 & 11\\
				5 		& 5.04 & 11.26 & 11 & 9 & 5.41 & 14.39 & 12 & -  & - & - & -\\
				\midrule
				\multicolumn{12}{c}{Spectral clustering}	\\
				2 		& 5.07 & 11.41 & 11 & 6 & 3.79 & 12.50 & 13 & 10 & 4.80 & 10.80 & 11\\
				\textbf{3} 		& \textbf{2.96} & \textbf{7.32} & \textbf{14} & 7 & 3.07 & 9.05 & 14 & 11 & 5.26 & 14.49 & 10\\
				4 		& 5.47 & 15.60 & 10 & 8 & 6.67 & 12.84 & 10 & 12 & 4.14 & 12.92 & 11\\
				5 		& 7.97 & 14.38 & 8 & 9 & 9.23 & 17.20 & 8 & -  & - & - & -\\
				\midrule
			\end{tabular} 
		}
	\end{table}
	
	The best result according to Table \ref{tab:5} was the $n_k$ equal to 3. There is a predilection for the spectral cluster initialization for straight line maneuvers actually, since the same type of result was observed for other maneuvers of the same type. As an example, Figure \ref{fig:10} shows two clusters in different extremes, one with high speed and another with a continuous breaking. 
	
	\subsection{Grouping trajectories in semantic significant behavior}
	\label{subsec:5.4}
	
	Assuming that the behavior profiles obtained via the clustering discussed are mostly influenced by the assertiveness of a road user and what interaction it describes, all clusters can be structured in function of one of these factors. For the turn left maneuver from Figure \ref{fig:1}, all the profiles obtained in Table \ref{tab:4} (30 clusters) are separated using the centroids obtained previously. 
	
	\begin{figure}[!h]
		\centering
		\begin{subfigure}[b]{\columnwidth}
			\centering
			\adjustbox{scale=0.8, trim=2mm 0mm 2mm 2mm, clip}{
				\includegraphics[width=1\textwidth]{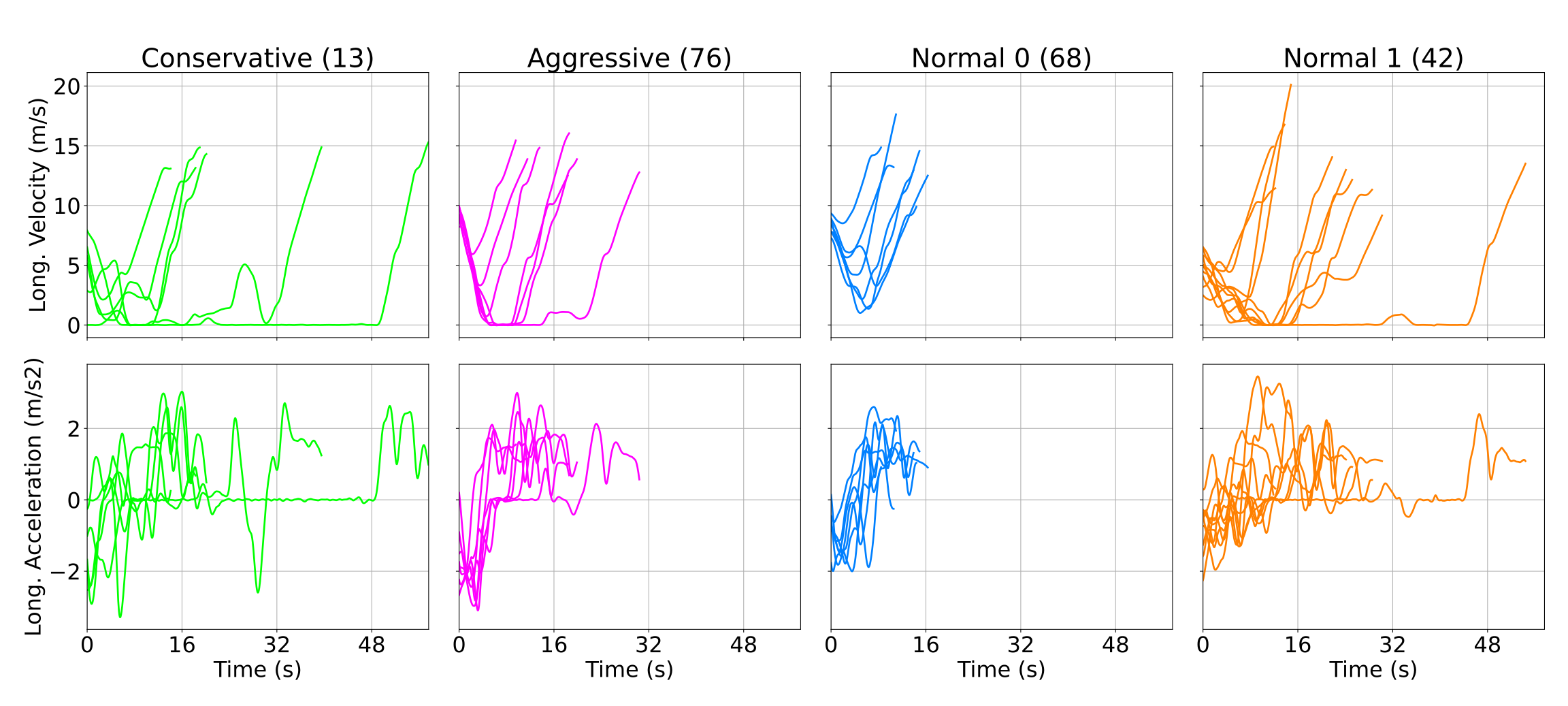} 
			}
			\caption{Assertiveness profiles}
			\label{fig:12a}
		\end{subfigure}
		\begin{subfigure}[b]{\columnwidth}
			\centering
			\adjustbox{scale=0.8, trim=2mm 0mm 2mm 2mm, clip}{
				\includegraphics[width=1\textwidth]{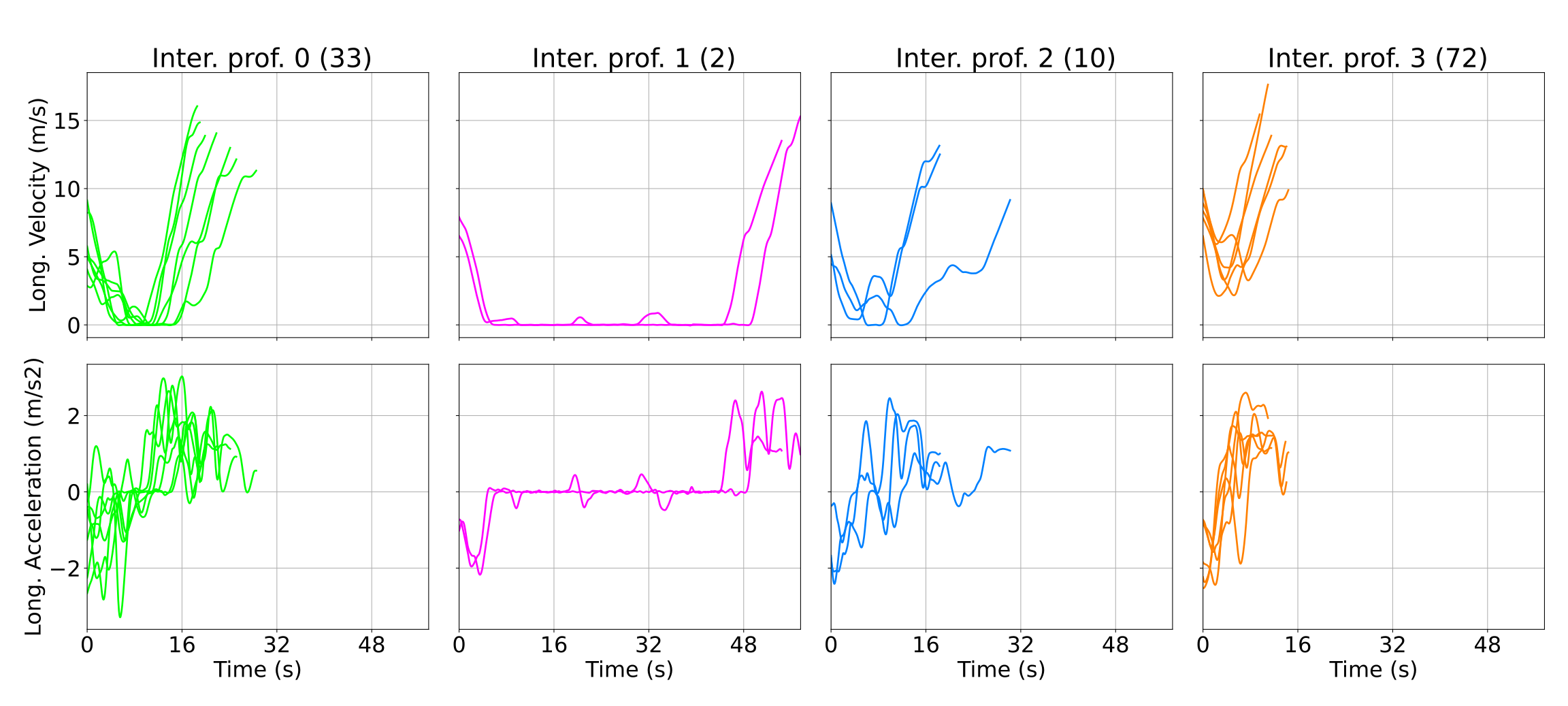} 
			}
		\end{subfigure}
		\begin{subfigure}[b]{\columnwidth}
			\centering
			\adjustbox{scale=0.8, trim=2mm 0mm 2mm 2mm, clip}{
				\includegraphics[width=1\textwidth]{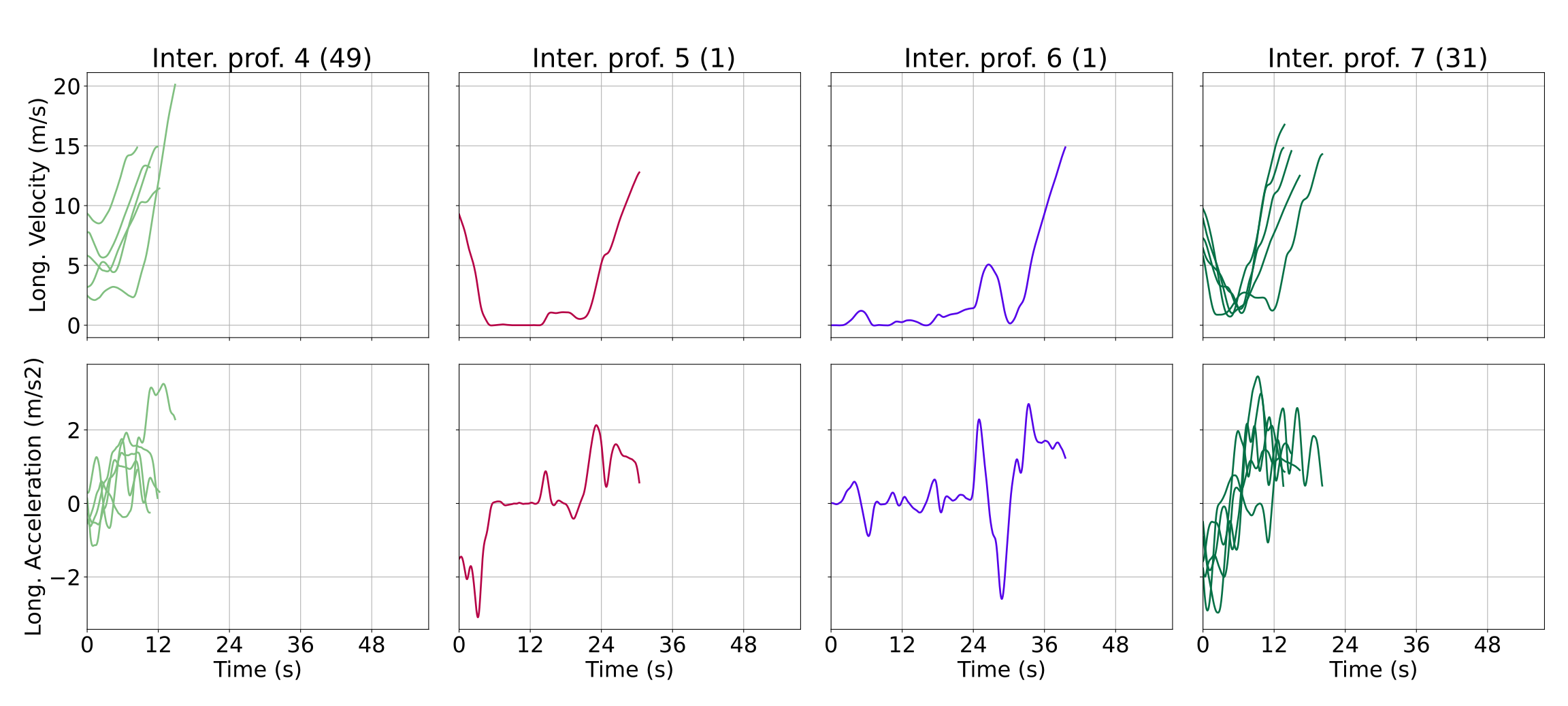} 
			}
			\caption{Interaction profiles}
			\label{fig:12b}
		\end{subfigure}
		\caption{Behavior profiles for turn right}
		\label{fig:12}
	\end{figure}
	
	Two clustering operations are executed with the centroids, with a fixed $n_k^{a}$ for the assertiveness and a certain number of interaction clusters, $n_k^{i}$, where $n_k^{a} \cdot n_k^{i} = n_k - c$ ($c$ adjusts the multiplication, all pairings are necessarily observed). There are 30 optimal clusters, defining 4 different assertiveness profiles and 8 interaction ones. The assertiveness clustering was done using three features: the initial longitudinal velocity, the period in witch the longitudinal acceleration stayed below -2 $m/s^2$, and lastly, above 1.5 $m/s^2$. These three features were normalized and clustered using the agglomeration algorithm, resulting in the Figure \ref{fig:12a}. 
	
\begin{figure}[!h]
		\centering
		\begin{subfigure}[b]{\columnwidth}
			\centering
			\adjustbox{scale=0.8, trim=2mm 0mm 2mm 2mm, clip}{
				\includegraphics[width=1\textwidth]{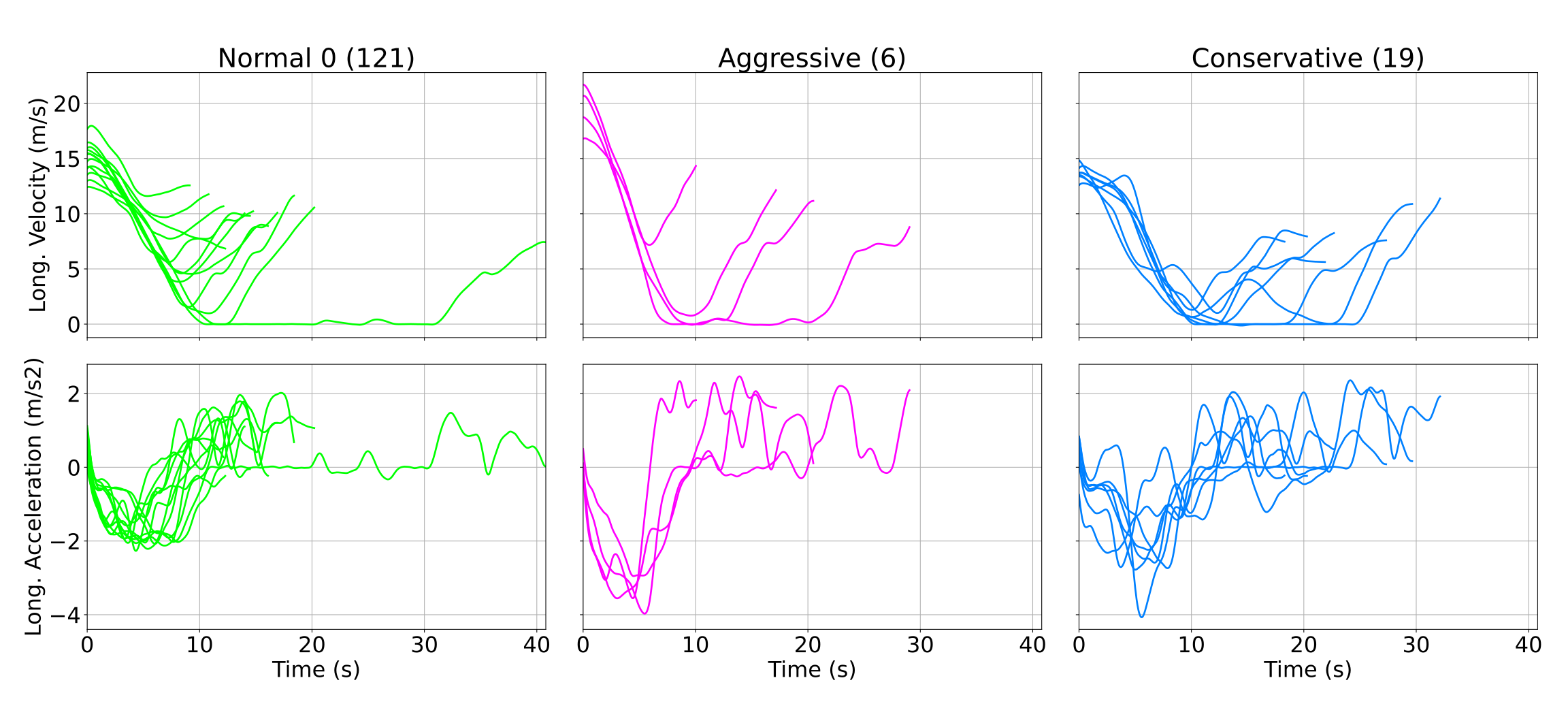} 
			}
		\end{subfigure}
		\caption{Assertiveness profiles for turn left}
		\label{fig:13}
	\end{figure}

	All clusters contain a range of different interactions, from slowing down to turn to stop altogether. As expected, the cluster that have a more aggressive result starts with a higher velocity and/or has a higher absolute acceleration while all behaviors on the conservative class got close to stopping or stopped before the intersection. Each name for the assertiveness profiles was given considering the average longitudinal velocity; thus, the order was aggressive, normal 0, normal 1 and conservative. For normal 0 no clusters that stop were detected, meaning that the data possibly did not capture all possible trajectories. As an example, for the turn left, all the behaviors were captured (Figure \ref{fig:13}, Table \ref{tab:4}).
	
	To obtain the interaction clusters in Figure \ref{fig:12b}, two features were used, in the same operation: a velocity reduction factor, the minimal velocity from each sample divided by its initial velocity, and the time below 1 $m/s$. The approach taken here to detect and catalog different interaction profiles concentrated itself on the ego-vehicle, without trying to explain every behavior, providing a wide range of possibilities. In Figure \ref{fig:12} there are multiple instances of interactions, from profiles 4, 3, 7 that reduce their speed to 2, 0, 1 that stopped completely during different times. This approach to obtain all modes of interaction for maneuvers is similar to \cite{samavi2023does}, discovering from data instead of from hand-crafted method as in \cite{gindele2010probabilistic}.
	
	\subsection{About initialization and convergence}
	\label{subsec:5.5}
	
	Throughout the procedure, two main points need to be commented: $t_{\tm{KL}}$ for each maneuver and the clustering convergence to a local minimum. Since the KL divergence comparison proposed in section \ref{sec:3} was done using the entire predicted state, i.e. equation \eqref{eq:24} and that each trajectory has a range of possible lanes to use, it does make sense to use different values of $t_{\tm{KL}}$ per maneuver. A simple exploration process was executed to discover the right interval for this constant. For the turn right it was 4 (Figure \ref{fig:1}), given the spread of possible positions at the trajectories' end, 3 for the turn left and 3.5 for the straight.
	
	Another important point in the clustering procedure is the method's convergence. According to the algorithm in Figure \ref{fig:11} the final result is obtained only after passing through the merge and split tests without any change in classification. The EM algorithm is guaranteed to converge, but not the split / merge transformation; sometimes the process also ends up in a loop, at which point the result is considered as final. However, given the procedure, only an asymptotic convergence to a local minimal can be assured. And even so, with the samples number increase comes a quadratic computational cost. For example, to cluster the turn right maneuver it took on average $2447.53 \pm 949.59$ seconds with $199$ trajectories, while for the straight maneuver it took $28999.2 \pm 8982.91$ seconds for 534 trajectories.  During the first iterations, the optimized variables $\mu_{nll}$ and $\sigma_{nll}$ decrease rapidly; afterwards the method keeps jumping from different $n_k$ values and obtaining only marginal gains. It could be beneficial to set a maximum number of iterations to stop the procedure before this variation. To mitigate this increase in complexity, during the centroid search in Figure \ref{fig:11} only the 30\% closest elements are considered as possible candidate for centroid, if the cluster has more than 100 samples.
	
	Finally, across the four initialization methods tested, there is no clear preference. The same procedure needs to be repeated with different $n_k^i$ values to obtain an optimal result, given that convergence to a global minimum, where the final values for $\mu_{nll}$, $\sigma_{nll}$ and $n_k$ would be always the same, was not observed in our tests. Even the metric being optimized, the sum $\mu_{ll}$ + $\sigma_{ll}$, is not ideal, since it is influenced by the increase on the number of clusters. The merge operation can counterbalance this effect, but it would be better to define a metric that also incorporates the number of clusters. 
	
	\section{CONCLUSIONS}
	
	The method proposed here uses the vehicles' dynamic characteristics to cluster multiple observations into different longitudinal behaviors, together with a probabilistic approach to determine the number of categories that exist. An implementation based on EKF and Mahalanobis distance for the clustering and the KL divergence for the determination on the number of clusters produced for three different maneuvers the distinct longitudinal behaviors expected. However, one drawback became evident from the obtained results: the convergence is dependent on $t_{\tm{KL}}$, which is itself unique for each maneuver and not easily obtained.
	
	Next steps involve using the obtained clusters to improve the prediction quality of some existent algorithm. Also, these longitudinal clusters will be used to train models for interactive independent agents, creating viable simulations for AV decision making evaluation. 
	
	\section*{Acknowledgment}
	
	The authors would like thanks Debora Barreto Ferreira for her invaluable assistance in refining the grammatical correctness and enhancing the language style of this paper.
	
	%
	%
	%
	
	
	
	\bibliographystyle{IEEEtran}
	\bibliography{iros_refs}

\begin{thebibliography}{10}
\providecommand{\url}[1]{#1}
\csname url@samestyle\endcsname
\providecommand{\newblock}{\relax}
\providecommand{\bibinfo}[2]{#2}
\providecommand{\BIBentrySTDinterwordspacing}{\spaceskip=0pt\relax}
\providecommand{\BIBentryALTinterwordstretchfactor}{4}
\providecommand{\BIBentryALTinterwordspacing}{\spaceskip=\fontdimen2\font plus
\BIBentryALTinterwordstretchfactor\fontdimen3\font minus
  \fontdimen4\font\relax}
\providecommand{\BIBforeignlanguage}[2]{{%
\expandafter\ifx\csname l@#1\endcsname\relax
\typeout{** WARNING: IEEEtran.bst: No hyphenation pattern has been}%
\typeout{** loaded for the language `#1'. Using the pattern for}%
\typeout{** the default language instead.}%
\else
\language=\csname l@#1\endcsname
\fi
#2}}
\providecommand{\BIBdecl}{\relax}
\BIBdecl

\bibitem{bahari2021injecting}
\BIBentryALTinterwordspacing
M.~Bahari, I.~Nejjar, and A.~Alahi, ``Injecting knowledge in data-driven
  vehicle trajectory predictors,'' \emph{Transportation Research Part C:
  Emerging Technologies}, vol. 128, p. 103010, 2021. [Online]. Available:
  \url{https://www.sciencedirect.com/science/article/pii/S0968090X21000425}
\BIBentrySTDinterwordspacing

\bibitem{hu2022review}
Z.~Hu, S.~Lou, Y.~Xing, X.~Wang, D.~Cao, and C.~Lv, ``Review and perspectives
  on driver digital twin and its enabling technologies for intelligent
  vehicles,'' \emph{IEEE Transactions on Intelligent Vehicles}, vol.~7, no.~3,
  pp. 417--440, 2022.

\bibitem{kuderer2015learning}
M.~Kuderer, S.~Gulati, and W.~Burgard, ``Learning driving styles for autonomous
  vehicles from demonstration,'' in \emph{2015 IEEE International Conference on
  Robotics and Automation (ICRA)}, 2015, pp. 2641--2646.

\bibitem{chai2019multipath}
\BIBentryALTinterwordspacing
Y.~Chai, B.~Sapp, M.~Bansal, and D.~Anguelov, ``Multipath: Multiple
  probabilistic anchor trajectory hypotheses for behavior prediction,'' 2019.
  [Online]. Available: \url{https://arxiv.org/abs/1910.05449}
\BIBentrySTDinterwordspacing

\bibitem{yanagisawa2006clustering}
Y.~Yanagisawa and T.~Satoh, ``Clustering multidimensional trajectories based on
  shape and velocity,'' in \emph{22nd International Conference on Data
  Engineering Workshops (ICDEW'06)}, 2006, pp. 12--12.

\bibitem{kerper2011driving}
M.~Kerper, C.~Wewetzer, H.~Trompeter, W.~Kiess, and M.~Mauve, ``Driving more
  efficiently - the use of inter-vehicle communication to predict a future
  velocity profile,'' in \emph{2011 IEEE 73rd Vehicular Technology Conference
  (VTC Spring)}, 2011, pp. 1--5.

\bibitem{demoura2023extraction}
N.~de~Moura and F.~Nashashibi, ``Extraction of vehicle behaviors at
  intersections,'' in \emph{2023 IEEE 26th International Conference on
  Intelligent Transportation Systems (ITSC)}, 2023, pp. 1779--1786.

\bibitem{lee2007trajectory}
\BIBentryALTinterwordspacing
J.-G. Lee, J.~Han, and K.-Y. Whang, ``Trajectory clustering: a
  partition-and-group framework,'' in \emph{Proceedings of the 2007 ACM SIGMOD
  International Conference on Management of Data}, ser. SIGMOD '07.\hskip 1em
  plus 0.5em minus 0.4em\relax New York, NY, USA: Association for Computing
  Machinery, 2007, p. 593–604. [Online]. Available:
  \url{https://doi.org/10.1145/1247480.1247546}
\BIBentrySTDinterwordspacing

\bibitem{yuan2017review}
\BIBentryALTinterwordspacing
G.~Yuan, P.~Sun, J.~Zhao, D.~Li, and C.~Wang, ``A review of moving object
  trajectory clustering algorithms,'' \emph{Artificial Intelligence Review},
  vol.~47, no.~1, pp. 123--144, Jan 2017. [Online]. Available:
  \url{https://doi.org/10.1007/s10462-016-9477-7}
\BIBentrySTDinterwordspacing

\bibitem{besse2016review}
P.~C. Besse, B.~Guillouet, J.-M. Loubes, and F.~Royer, ``Review and perspective
  for distance-based clustering of vehicle trajectories,'' \emph{IEEE
  Transactions on Intelligent Transportation Systems}, vol.~17, no.~11, pp.
  3306--3317, 2016.

\bibitem{martinsson2018clustering}
J.~Martinsson, N.~Mohammadiha, and A.~Schliep, ``Clustering vehicle maneuver
  trajectories using mixtures of hidden markov models,'' in \emph{2018 21st
  International Conference on Intelligent Transportation Systems (ITSC)}, 2018,
  pp. 3698--3705.

\bibitem{dewei2019trajectory}
D.~Yi, J.~Su, C.~Liu, and W.-H. Chen, ``Trajectory clustering aided
  personalized driver intention prediction for intelligent vehicles,''
  \emph{IEEE Transactions on Industrial Informatics}, vol.~15, no.~6, pp.
  3693--3702, 2019.

\bibitem{sung2012traj}
C.~Sung, D.~Feldman, and D.~Rus, ``Trajectory clustering for motion
  prediction,'' in \emph{2012 IEEE/RSJ International Conference on Intelligent
  Robots and Systems}, 2012, pp. 1547--1552.

\bibitem{deo2018how}
N.~Deo, A.~Rangesh, and M.~M. Trivedi, ``How would surround vehicles move? a
  unified framework for maneuver classification and motion prediction,''
  \emph{IEEE Transactions on Intelligent Vehicles}, vol.~3, no.~2, pp.
  129--140, 2018.

\bibitem{li2018depp}
X.~Li, K.~Zhao, G.~Cong, C.~S. Jensen, and W.~Wei, ``Deep representation
  learning for trajectory similarity computation,'' in \emph{2018 IEEE 34th
  International Conference on Data Engineering (ICDE)}, 2018, pp. 617--628.

\bibitem{fang2021e2dtc}
Z.~Fang, Y.~Du, L.~Chen, Y.~Hu, Y.~Gao, and G.~Chen, ``E2dtc: An end to end
  deep trajectory clustering framework via self-training,'' in \emph{2021 IEEE
  37th International Conference on Data Engineering (ICDE)}, 2021, pp.
  696--707.

\bibitem{sutskever2014sequence}
\BIBentryALTinterwordspacing
I.~Sutskever, O.~Vinyals, and Q.~V. Le, ``Sequence to sequence learning with
  neural networks,'' 2014. [Online]. Available:
  \url{https://arxiv.org/abs/1409.3215}
\BIBentrySTDinterwordspacing

\bibitem{harmening2020deep}
\BIBentryALTinterwordspacing
N.~Harmening, M.~Biloš, and S.~Günnemann, ``Deep representation learning and
  clustering of traffic scenarios,'' 2020. [Online]. Available:
  \url{https://arxiv.org/abs/2007.07740}
\BIBentrySTDinterwordspacing

\bibitem{shouno2018deep}
O.~Shouno, ``Deep unsupervised learning of a topological map of vehicle
  maneuvers for characterizing driving styles,'' in \emph{2018 21st
  International Conference on Intelligent Transportation Systems (ITSC)}, 2018,
  pp. 2917--2922.

\bibitem{bock2020}
J.~Bock, R.~Krajewski, T.~Moers, S.~Runde, L.~Vater, and L.~Eckstein, ``The ind
  dataset: A drone dataset of naturalistic road user trajectories at german
  intersections,'' in \emph{2020 IEEE Intelligent Vehicles Symposium
  (IV)}.\hskip 1em plus 0.5em minus 0.4em\relax IEEE, 2020, pp. 1929--1934.

\bibitem{demoura2024hal}
\BIBentryALTinterwordspacing
N.~de~Moura, A.~Gervreau-Mercier, F.~Garrido, and F.~Nashashibi, ``Fast
  maneuver recovery from aerial observation: trajectory clustering and outliers
  rejection,'' in \emph{2024 IEEE Intelligent Vehicles Symposium (IV)}, 2024.
  [Online]. Available: \url{https://inria.hal.science/hal-04497008}
\BIBentrySTDinterwordspacing

\bibitem{thrun2005probabilistic}
S.~Thrun, W.~Burgard, and D.~Fox, \emph{Probabilistic Robotics (Intelligent
  Robotics and Autonomous Agents)}.\hskip 1em plus 0.5em minus 0.4em\relax The
  MIT Press, 2005.

\bibitem{krener2002convergence}
A.~J. Krener, ``The convergence of the extended kalman filter,'' 2002.

\bibitem{demoura2021governing}
\BIBentryALTinterwordspacing
N.~D. Moura Martins~Gomes, ``{Governing Automated Vehicle Behavior},'' Theses,
  {Sorbonne Universit{\'e}}, Jun. 2021. [Online]. Available:
  \url{https://theses.hal.science/tel-03709568}
\BIBentrySTDinterwordspacing

\bibitem{hoffmann2007autonomous}
G.~M. Hoffmann, C.~J. Tomlin, M.~Montemerlo, and S.~Thrun, ``Autonomous
  automobile trajectory tracking for off-road driving: Controller design,
  experimental validation and racing,'' in \emph{2007 American Control
  Conference}, 2007, pp. 2296--2301.

\bibitem{hashemi2019generalized}
N.~Hashemi and J.~Ruths, ``Generalized chi-squared detector for lti systems
  with non-gaussian noise,'' in \emph{2019 American Control Conference (ACC)},
  2019, pp. 404--410.

\bibitem{rodriguez2010visual}
S.~A. Rodríguez~F, V.~Frémont, P.~Bonnifait, and V.~Cherfaoui, ``Visual
  confirmation of mobile objects tracked by a multi-layer lidar,'' in
  \emph{13th International IEEE Conference on Intelligent Transportation
  Systems}, 2010, pp. 849--854.

\bibitem{chang2014robust}
\BIBentryALTinterwordspacing
G.~Chang, ``Robust kalman filtering based on mahalanobis distance as outlier
  judging criterion,'' \emph{Journal of Geodesy}, vol.~88, no.~4, pp. 391--401,
  Apr 2014. [Online]. Available:
  \url{https://doi.org/10.1007/s00190-013-0690-8}
\BIBentrySTDinterwordspacing

\bibitem{mclachlan2007algorithm}
G.~J. McLachlan and T.~Krishnan, \emph{The EM algorithm and extensions}.\hskip
  1em plus 0.5em minus 0.4em\relax John Wiley \& Sons, 2007.

\bibitem{saraiva2019data}
\BIBentryALTinterwordspacing
E.~F. Saraiva, C.~Pereira, and A.~K. Suzuki, ``A data-driven selection of the
  number of clusters in the dirichlet allocation model via bayesian mixture
  modelling,'' \emph{Journal of Statistical Computation and Simulation},
  vol.~89, no.~15, pp. 2848--2870, 2019. [Online]. Available:
  \url{https://doi.org/10.1080/00949655.2019.1643345}
\BIBentrySTDinterwordspacing

\bibitem{samavi2023does}
S.~Samavi, F.~Shkurti, and A.~P. Schoellig, ``Does unpredictability influence
  driving behavior?'' in \emph{2023 IEEE/RSJ International Conference on
  Intelligent Robots and Systems (IROS)}, 2023, pp. 1720--1727.

\bibitem{gindele2010probabilistic}
T.~Gindele, S.~Brechtel, and R.~Dillmann, ``A probabilistic model for
  estimating driver behaviors and vehicle trajectories in traffic
  environments,'' in \emph{13th International IEEE Conference on Intelligent
  Transportation Systems}, 2010, pp. 1625--1631.

\end{thebibliography}
	
\end{document}